\documentclass{article}
\PassOptionsToPackage{numbers, compress}{natbib}
\usepackage[preprint]{neurips_2026}

\usepackage[utf8]{inputenc} 
\usepackage[T1]{fontenc}    
\usepackage{hyperref}       
\usepackage{url}            
\usepackage{booktabs}       
\usepackage{amsfonts}       
\usepackage{nicefrac}       
\usepackage{microtype}      
\usepackage[table]{xcolor}
\usepackage{amsmath}
\usepackage{amssymb}
\usepackage{multirow}
\usepackage{graphicx}
\usepackage[many]{tcolorbox}
\usepackage{fancyvrb}
\usepackage{subcaption}
\usepackage[export]{adjustbox}
\usepackage{caption}
\usepackage{wrapfig}
\usepackage{mathtools}
\usepackage[titletoc,title]{appendix}
\usepackage{titletoc}
\usepackage{enumitem}
\usepackage{makecell}
\usepackage{float}
\usepackage{fontawesome5}
\usepackage{placeins}

\colorlet{takeawaybg}{blue!10}
\definecolor{grey}{rgb}{0.9,0.9,0.9}
\newtcolorbox{takeawaysbox}{
    enhanced,
    colback=takeawaybg,
    colframe=black,
    boxrule=1.0pt,
    arc=6pt, 
    auto outer arc,
    top=8pt,
    bottom=4pt,
    left=6pt,
    right=6pt,
    fontupper=\normalsize,
    title=Takeaways,
    coltitle=black,
    fonttitle=\bfseries,
    attach boxed title to top left={xshift=12pt, yshift=-8pt},
    boxed title style={
        colback=grey,
        colframe=black,
        size=small,
        arc=3pt, 
        bottom=0.5pt, top=0.5pt, left=2pt, right=2pt
    }
}

\title{Learning What to Imitate: \\ Entropy-Aware Distribution Mixing}

\author{%
\begin{tabular}{c}
Juan Garcia Giraldo\textsuperscript{1,2} \qquad
Matteo Santelmo\textsuperscript{1,2} \qquad
Eduard Durech\textsuperscript{2} \\
Imanol Schlag\textsuperscript{2} \qquad
Valentina Pyatkin\textsuperscript{2,3} \qquad
Antoine Bosselut\textsuperscript{1} \\[0.6em]
\textnormal{%
\textsuperscript{1}EPFL \qquad
\textsuperscript{2}ETH AI Center \qquad
\textsuperscript{3}Allen Institute for AI
} \\[0.5em]
\textnormal{\texttt{Corresponding author: juangarciagiraldo@gmail.com}}
\end{tabular}
}

\begin{document}

\maketitle

\vspace{-2.5em}
\begin{center}
    {\faGithub\ \href{https://github.com/d23845jg/entropy-aware-mixing}{\texttt{d23845jg/entropy-aware-mixing}}}
\end{center}

\begin{abstract}
Small language models are often post-trained as students on reasoning traces from stronger teacher models to efficiently learn new skills. However, token-level imitation on traces that lie far outside the student’s expected distribution often produces \textit{confident conflicts}, whereby the student is required to imitate a continuation that it deems unlikely (i.e., low-probability) despite being confident in a different continuation (i.e., in a low-entropy state). To mitigate the degradation in generalisation and catastrophic forgetting caused by these conflicts, we propose \textbf{Entropy-Aware Mixing}: a dynamic per-token interpolation of the student and teacher distributions, gated by the student's predictive entropy. We implement both convex and geometric interpolations for both offline trace generation (via speculative decoding, then SFT) and on-policy forward-KL distillation. Our results show that entropy-aware mixing stabilises distillation, improving in-distribution and out-of-distribution math reasoning while better preserving general capabilities than fixed-teacher supervision. Nonetheless, the optimal entropy schedule depends on the training source, with offline-generated traces favouring concave schedules (greater overall teacher influence) and on-policy training favouring linear or convex schedules (teacher concentrated in high-entropy states).
\end{abstract}

\vspace{-0.7em}
\section{Introduction}
Post-training small language models for specialised domains often degrades their existing capabilities on more general tasks \citep{lin2026sftdoesnthurtgeneral,luo2025empiricalstudycatastrophicforgetting,shi2024continuallearninglargelanguage}. This effect is even more pronounced when the training signal is distilled from a stronger teacher \citep{agarwal2024policy,chen2025retainingdoingroleonpolicy,minillm,kim2025wordsreasoningtracestailored,li2026rethinkingonpolicydistillationlarge,lu2025onpolicydistillation}. When training on teacher-generated traces, small models efficiently learn important semantic signals, such as decomposition choices and reasoning fragments, but they also train on token-level artefacts such as notation or formatting choices that undermine learning. Consequently, even when the teacher produces a correct solution, token-level imitation of these traces forces the student to train on continuations far outside its learned expected distribution (Section~\ref{sec:confident_conflicts}), a friction compounded by the fact that fine-tuning and distillation objectives apply supervision uniformly across all steps. This student-teacher expectation divergence may limit the effective use of existing representations of the student, and the resulting model updates may degrade capabilities on tasks unrelated to the fine-tuning domain. 

While recent methods mitigate the impact of this divergence by down-weighting distillation losses when student confidence is high \citep{diao2026entropy,lin2026sftdoesnthurtgeneral,wu2026generalisationsftreinforcementlearning,zhang2026onpolicysftdistributiondiscriminant}, continued optimisation against off-policy teacher traces maintains the underlying issue. 
In this work, rather than changing \textit{how strongly} the student learns from static teacher traces, we create a target distribution that produces more student-aligned traces by adaptively mixing the student and teacher distributions.

To this end, we propose \textbf{Entropy-Aware Mixing}, a token-level framework that dynamically combines student and teacher distributions using the student's predictive entropy. Unlike fixed interpolation coefficients, the induced mixture remains close to the student in low-entropy states, preserving high-confidence behaviour, and shifts towards the teacher in uncertain states, concentrating supervision where it is most useful. We study both convex and geometric mixtures, which capture student-teacher agreement in different ways, and apply them in two regimes: offline trace generation followed by SFT, and online distillation under forward KL. Experiments across multiple model families (Qwen~\citep{qwen3}, Liquid AI~\citep{liquidAI2026350M}, and Granite ~\citep{ibm_granite_33}), student scales (305M-2B) and reasoning domains (math~\citep{hendrycks2021math}, code generation~\citep{xia2025leetcodedatasettemporaldatasetrobust}) demonstrate that entropy-aware mixing improves reasoning and better preserves generalisation compared to fixed-teacher supervision.

We summarize our contributions as follows:
\begin{enumerate}
    \item We identify \emph{confident conflicts}, low-probability tokens in low-entropy student states, as the source of destructive supervision in post-training, and show they are not limited to final answers but appear throughout reasoning traces.
    \item We propose entropy-aware mixed targets, a target-shaping alternative to loss reweighting, in both convex (\emph{ConMix}) and geometric (\emph{GeoMix}) forms, and instantiate them in offline (speculative-decoding followed by SFT) and online (on-policy forward-KL) distillation regimes.
    \item We demonstrate that entropy-aware method delivers strong in-distribution and out-of-distribution mathematical performance. Additional experiments show that these gains generalise across student sizes, domains, model families, and mixture definitions.
\end{enumerate}

\begin{figure}[!tbp]
    \includegraphics[width=\textwidth]{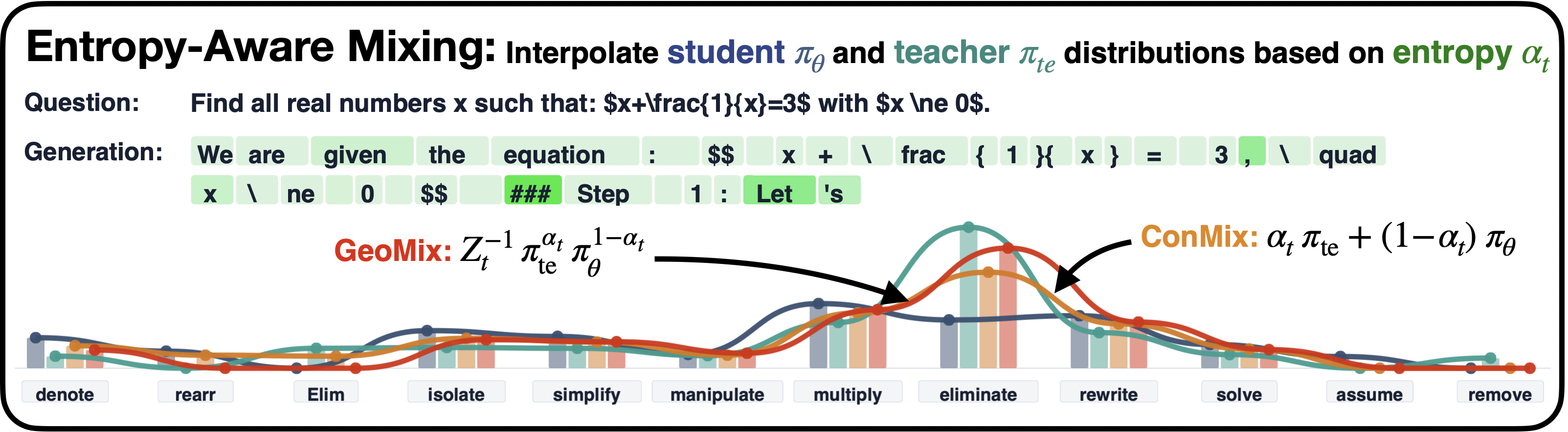}
    \captionof{figure}{Conceptual illustration of Entropy-Aware Mixing. The entropy-based interpolation coefficient $\alpha_t$ controls the shift from $\pi_\theta$ toward $\pi_{\mathrm{te}}$ via either convex (\emph{ConMix}) or geometric (\emph{GeoMix}) mixtures.}
\end{figure}

\section{Background} \label{sec:confident_conflicts}
A key challenge in post-training is that the supervision signal is often \emph{off-policy} for the student, particularly when derived from a fixed dataset of teacher-generated traces. While these traces express learnable reasoning paths to final answers, they also reflect teacher-specific artefacts of problem decomposition, notation, formatting, and local arithmetic steps. Consequently, strict token-level imitation of these reasoning traces can force the student to learn token-level patterns far from its expected distribution, undermining learning and generally deteriorating other capabilities. These updates are most disruptive when the student is already confident in a different token (\textit{i.e.}, in a low-entropy state), causing it to substantially overwrite learnt representations.

\begin{figure}[!tbp]
    \centering
    \begin{subfigure}[t]{0.60\textwidth}
        \centering
        \includegraphics[width=\textwidth, valign=t]{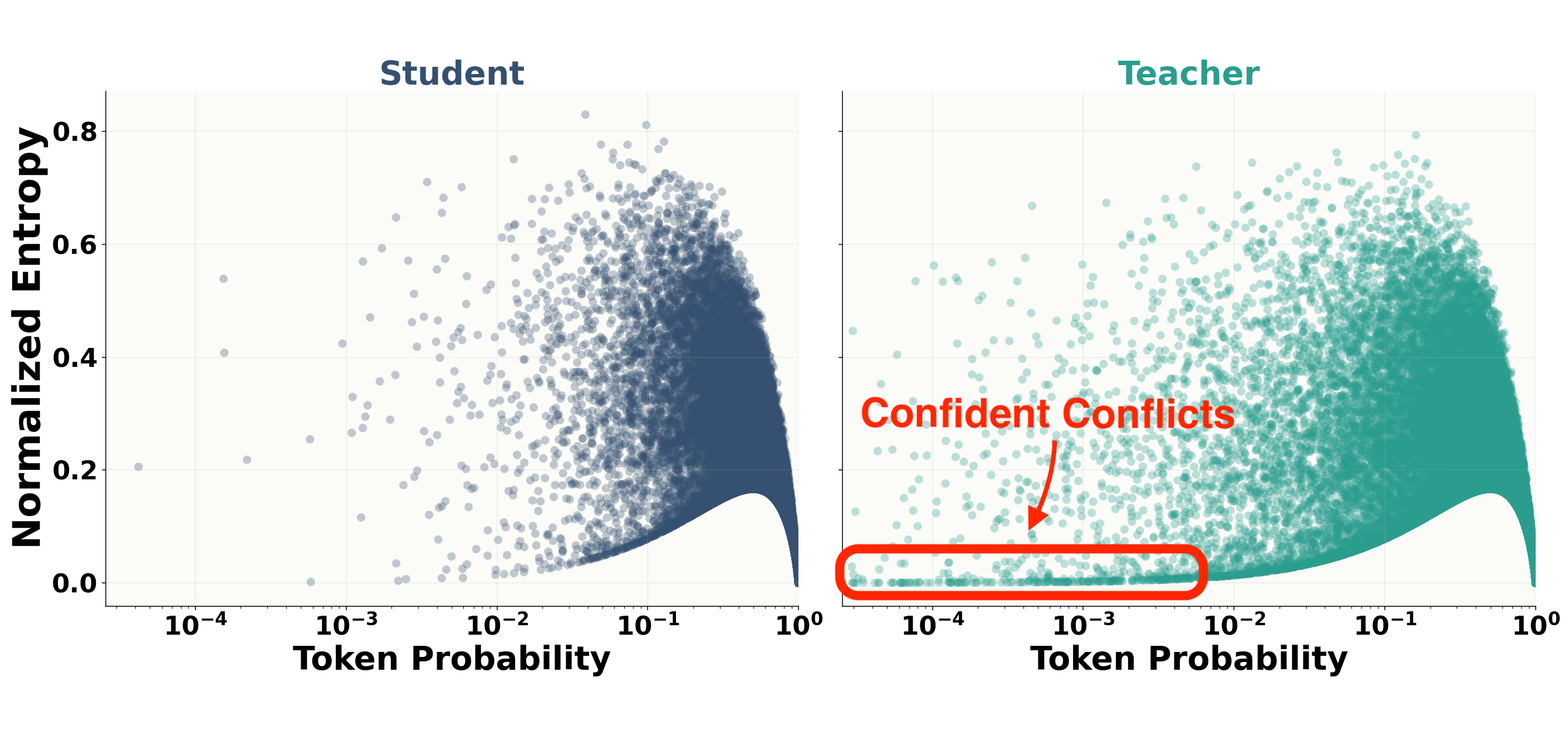}
        \caption{Normalised student entropy vs token probability on student-generated (left) and teacher-generated trajectories (right).}
        \label{fig:entropy_prob}
    \end{subfigure}
    \hfill
    \begin{subfigure}[t]{0.36\textwidth}
        \centering
        \includegraphics[width=\textwidth, valign=t]{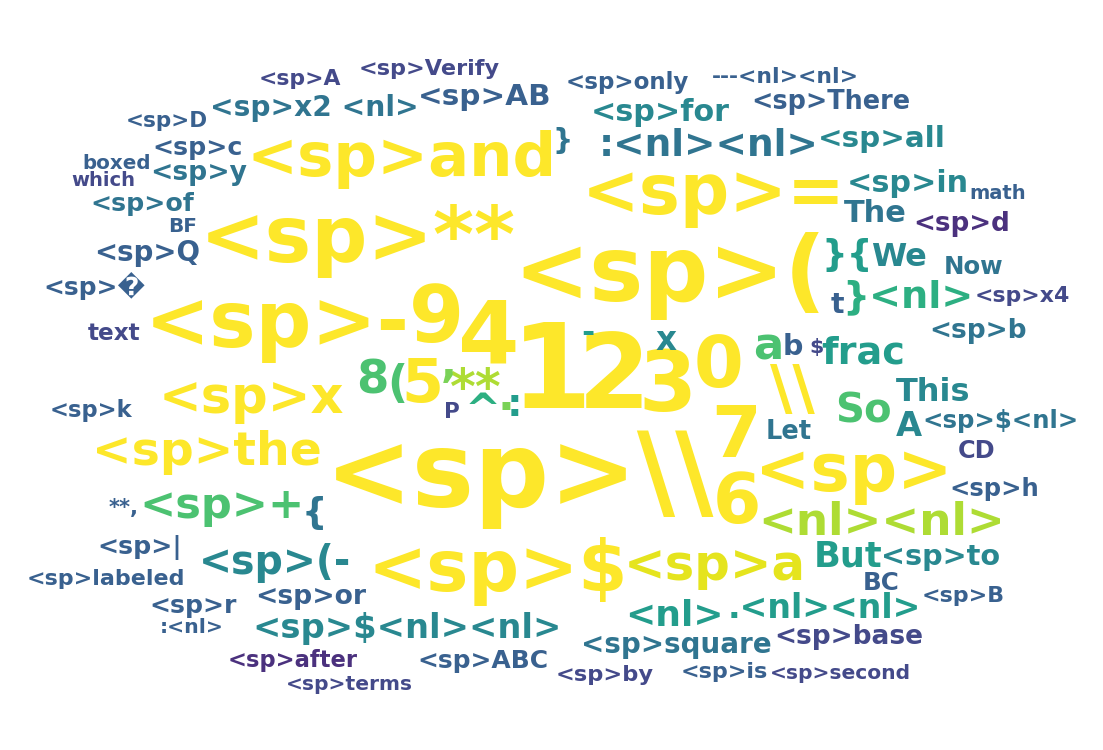}
        \caption{Confident conflicts are dominated by arithmetic fragments, whitespace, formatting markers, and \LaTeX{} symbols.}
        \label{fig:word_cloud}
    \end{subfigure}
    \caption{Reasoning-trace analysis with Qwen3-1.7B student and a Qwen3-8B teacher on a subset of MATH (thinking mode disabled, temperature \(1.0\), top-\(p=1.0\)). Teacher trajectories expose the student to many confident conflicts; student trajectories largely suppress them.}
    \label{fig:confident_conflicts}
\end{figure}

Following \citet{diao2026entropy}, we refer to these states as \emph{confident conflicts}: states in which the student has low predictive entropy yet assigns low probability to the target token. Figure~\ref{fig:entropy_prob} shows that such divergent targets are common when training on teacher-generated trajectories, even when teacher and student come from the same model family. By contrast, they are largely uncommon in student-generated trajectories, which remain closer to the student's own distribution. Figure~\ref{fig:word_cloud} further shows that these conflicts are not limited to final answers. In mathematical reasoning traces, they frequently involve boilerplate tokens such as whitespace (\texttt{<sp>}), newline character (\texttt{<nl>}), \LaTeX{} symbols such as \texttt{\$} and \texttt{\textbackslash frac}, digits, and short numerical fragments.

\citet{diao2026entropy} mitigated confident conflicts by reweighting the SFT loss using the student's entropy, reducing the influence of low-entropy states where the student could confidently disagree with the target. However, selectively down-weighting tokens on teacher-generated traces can amplify the mismatch between teacher-forced training contexts and the student-generated contexts encountered at inference. Instead, recent mixture-based objectives~\citep{kodistillm,shin2026amid} interpolate the student-teacher distributions, with prior work empirically favouring interpolation coefficients that assign relatively little weight to the teacher at the expense of limiting the ability to inject stronger guidance to the student.


In the following sections, we show that supervision is more effective when the target distribution is adaptively shaped around the student through our entropy-aware interpolation mechanism, eliminating the need to mask confident conflicts from off-policy teacher trajectories or rely on small interpolation coefficients that apply guidance uniformly.

\section{Entropy-Aware Mixing} \label{sec:entropy_aware_mixing}
Existing distillation objectives typically use a fixed target distribution, treating every token in the teacher trace as an effective ground truth. However, in confident conflicts, blindly matching the teacher distribution hinders generalisation by forcing the student to learn from targets that are poorly aligned with its existing representations \citep{diao2026entropy,zhang2026onpolicysftdistributiondiscriminant}. Motivated by this issue, we replace the static target distribution with an adaptive mixture that interpolates between the student–teacher distributions according to the student’s entropy.

\subsection{Defining the Mixture-Based Distribution}
We seek a target distribution that remains close to the student when it is confident, yet shifts towards the teacher when the student is uncertain. As such, we formulate two mixtures\footnote{We note that AMiD~\citep{shin2026amid} covers both mixture forms within a family of mixtures they taxonomize. We discuss our work in the context of theirs in Appendix~\ref{app:amid}.} of the student and teacher distributions, \textit{ConMix}, which linearly interpolates the probability distributions:

\begin{equation}
    \pi_{\mathrm{mix}}(y_t \mid \mathbf{c}_t) = (1-\alpha_t)\pi_{\theta}(y_t \mid \mathbf{c}_t) + \alpha_t \pi_{\mathrm{te}}(y_t \mid \mathbf{c}_t)
\end{equation}

and \textit{GeoMix}, which computes a geometric average of both distributions, more heavily favouring tokens both models consider likely:

\begin{equation}
    \pi_{\mathrm{mix}}(y_t \mid \mathbf{c}_t) = \frac{1}{Z_t} \pi_{\theta}(y_t \mid \mathbf{c}_t)^{1-\alpha_t} \pi_{\mathrm{te}}(y_t \mid \mathbf{c}_t)^{\alpha_t}
\end{equation}

In both mixtures, $\mathbf{c}_t = (\mathbf{x}, \mathbf{y}_{<t})$ is the concatenation of the prompt and previous response tokens at each decoding step $t$. In \textit{GeoMix}, \(Z_t=\sum_{y'} \pi_{\mathrm{te}}(y' \mid \mathbf{c}_t)^{\alpha_t}\pi_{\theta}(y' \mid \mathbf{c}_t)^{1-\alpha_t}\) is the normalisation constant required to ensure that the geometric mixture defines a valid probability distribution. In both formulations, the interpolation coefficient $\alpha_t$ controls the degree of teacher influence at each decoding step $t$. \(\alpha_t \approx 0\) recovers the student distribution $\pi_{\theta}$, whereas \(\alpha_t \approx 1\) recovers the teacher distribution $\pi_{\mathrm{te}}$, thereby controlling when teacher guidance is applied.

\subsection{Entropy-Based Interpolation Schedule} \label{sec:adaptive_signal}
To mitigate the confident conflicts described in Section~\ref{sec:confident_conflicts}, we employ an uncertainty-aware interpolation strategy. At each decoding step, we approximate the Shannon entropy \citep{shannon1948mathematical} of the student over the top-$K$ tokens to determine the interpolation coefficient $\alpha_t$, preserving the student’s distribution in low-entropy regimes.

Let \(\mathcal{V}_t^K\) denote the set of the \(K\) most probable tokens under \(\pi_\theta(\cdot \mid \mathbf{c}_t)\). Given that \(P_t(\cdot)\) denotes the student distribution renormalised over this subset, we define the top-\(K\) entropy as:

\begin{align*}
    H_t^{\mathrm{top}\text{-}K}
    &= - \sum_{y \in \mathcal{V}_t^K} P_t(y)\log P_t(y), \\
    \tilde{H}_t
    &= \frac{H_t^{\mathrm{top}\text{-}K}}{\log K}.
\end{align*}
The normalised entropy \(\tilde{H}_t \in [0,1]\) serves as our token-level uncertainty signal. We choose \(K\) empirically so that this top-\(K\) estimate remains strongly correlated with full-vocabulary entropy while avoiding full-vocabulary computation. Finally, we map \(\tilde{H}_t\) to a mixing coefficient \(\alpha_t = f(\tilde{H}_t)\), where \(f:[0,1]\!\to\![0,1]\) is a monotonic transformation that controls the interpolation strength under the proposed mixture targets. The choice of \(f\) determines how the mixture varies with entropy: convex functions, such as \(f(x)=x^2\) (\texttt{sq}), concentrate teacher guidance in high-entropy states, whereas concave functions, like \(f(x)=x^{1/2}\) (\texttt{sqrt}), assign more relative guidance to low-entropy states.

\subsection{Training Objectives} \label{sec:training_obj}
We evaluate $\pi_{\mathrm{mix}}$ under two training objectives: (i) in the \emph{offline} setting, we use $\pi_{\mathrm{mix}}$ to generate static reasoning traces for SFT; (ii) in the \emph{online} setting, $\pi_{\mathrm{mix}}$ is the per-token target in a forward-KL objective.

\paragraph{Offline Distillation: Synthetic Trace Generation for Supervised Fine-Tuning.} 
We construct a fixed dataset \(\mathcal{D}\) of reasoning traces by sampling from the entropy-aware target \(\pi_{\mathrm{mix}}\), and then train the student with standard supervised fine-tuning. To efficiently generate large amounts of data, we adopt speculative decoding (SD) \citep{chen2023accelerating,leviathan2023fast}, using the student model \(\pi_\theta\) as the draft model and the mixed distribution \(\pi_{\mathrm{mix}}\) as the target distribution. This approach preserves the exact marginal target distribution while improving decoding efficiency. Sampling from the convex and geometric mixtures yields two distinct offline datasets, \textcolor{blue!50}{\emph{\textbf{ConMix-SFT}}} and \textcolor{blue!50}{\emph{\textbf{GeoMix-SFT}}}. 


\paragraph{Online Distillation: On-Policy Forward-KL.} 
In the online setting, we directly optimise towards the proposed mixture by optimising the on-policy forward-KL (FKL) objective \(D_{\mathrm{KL}}\!\left(\pi_{\mathrm{mix}}(\cdot \mid \mathbf{c}_t)\,\|\, \pi_\theta(\cdot \mid \mathbf{c}_t)\right)\) using student-generated samples (Appendix~\ref{app:background}). This yields two variants, \textcolor{blue!50}{\emph{\textbf{ConMix-FKL}}} and \textcolor{blue!50}{\emph{\textbf{GeoMix-FKL}}}, based on the convex and geometric mixtures, respectively. In practice, we construct the mixture using the renormalised top-$K$ teacher distribution (Appendix~\ref{app:hyper_params}) and apply the stop-gradient operator $\operatorname{sg}[\cdot]$ to $\pi_{\mathrm{mix}}$, treating it as fixed during training due to its dependence on $\pi_{\theta}$. 

\section{Experimental Setup} \label{sec:experiment}
As defined in Section~\ref{sec:training_obj}, we evaluate our method in two settings: (a) supervised fine-tuning on traces sampled from our entropy-aware mixture, and (b) online distillation. We structure the results around mathematical reasoning performance, general capabilities, token-level entropy dynamics, and the effect of the entropy transformation $f$ on downstream performance.

\paragraph{Models.}
We focus our experiments on the Qwen3 family~\citep{qwen3}, using Qwen3-8B non-thinking as the teacher and its smaller variants as students. For offline distillation, we employ Qwen3-0.6B non-thinking and Qwen3-1.7B non-thinking to allow direct comparisons against self-distilled traces, whereas we adopt the corresponding base models for on-policy distillation. Due to space constraints, we report results using the 1.7B student in Section~\ref{sec:math_performance}, and defer comprehensive 0.6B results to Appendix~\ref{app:qwen3_results}. Additionally, we provide supplementary results using LFM2.5-350M~\citep{liquidai2025lfm2} and Granite-3.2-2B-Instruct~\citep{ibm_granite_33} as student models (with larger models from the same families as teacher models) in Appendix~\ref{app:extra_models}.

\paragraph{Training.}
Unless otherwise specified, the following configurations apply during training. All training is conducted on mathematical reasoning problems from the MATH~\citep{hendrycks2021math} training split. In the offline setting, we construct a static dataset by sampling eight rollouts per question and retaining one of them. When more than one sampled rollout is correct, we select the correct rollout uniformly at random; otherwise, we select one incorrect rollout uniformly at random. For the online setting, training is fully on-policy: at each optimisation step, we sample one rollout per question from the current student policy. All generations use temperature $1.0$, top-$p=1.0$, and a maximum response length of 4096 tokens. 

The entropy-based interpolation coefficient $\alpha_t$ in our mixtures is computed by applying a transformation function $f$ to the normalised student entropy. Throughout the experiments, we use the concave transformation $f(x)=x^{\frac{1}{4}}$ (\texttt{sqrt2}) for \emph{offline} trace generation and the convex transformation $f(x)=x^{2}$ (\texttt{sq}) for \emph{online} forward-KL distillation. Section~\ref{sec:entropy_transform} presents an ablation of alternative entropy transformations.

Offline learning uses vLLM~\citep{kwon2023efficient} for speculative-decoding trace generation and \texttt{verl}~\citep{sheng2024hybridflow} for supervised fine-tuning. Online learning is implemented entirely in \texttt{verl}. We do not perform an exhaustive hyperparameter search for each baseline and instead follow a setup closely aligned with \citet{jin2026entropy}. Full hyperparameters and training details are deferred to Appendix~\ref{app:hyper_params}. All data generation, training, and evaluation are performed across three random seeds; we report the mean and standard deviation. Each run uses a single node with four NVIDIA GH200 120GB GPUs.

\paragraph{Evaluation.}
We evaluate our approach across three domains: (1) \emph{in-distribution mathematical reasoning} on MATH500~\citep{hendrycks2021math} and GSM8K~\citep{cobbe2021gsm8k}; (2) \emph{out-of-distribution (OOD) math generalisation} on AIME24/25~\citep{maa_aime}, AMC23~\citep{maa_amc8_2023}, and Minerva~\citep{lewkowycz2022minerva}; and (3) \emph{general capabilities} via HellaSwag~\citep{zellers2019hellaswag} (length-normalised accuracy), IFEval~\citep{zhou2023ifeval} (prompt-level strict accuracy), and generative MMLU~\citep{hendrycks2021mmlu} (accuracy). For all math tasks, we sample 64 responses per question, reporting Avg@64, Pass@8, and Pass@64. For general capabilities, we report Avg@8. Unless specified otherwise, decoding uses temperature $T=1.0$, top-$p=0.8$, and a maximum length of 8192 tokens. Due to space constraints, we report a subset of benchmarks below and provide full evaluations in Appendix~\ref{app:qwen3_results}.

\paragraph{Baselines.}
We evaluate against standard baselines in both the offline setting (SFT on static reasoning traces) and the online setting (on-policy distillation). In the \emph{offline} setting, we compare against SFT on: (1) student-generated traces, (2) teacher-generated traces, and (3) SKD-inspired sampling~\citep{xu2024speculative}. We further evaluate (4) Entropy-Adaptive Fine-Tuning (EAFT)~\citep{diao2026entropy} on teacher traces, which applies a student-entropy penalty to the cross-entropy loss to suppress updates on low-entropy states. In the \emph{online} setting, we compare with: (1) standard forward-KL distillation, (2) OPD with PPO-style importance sampling and clipping~\citep{lu2025onpolicydistillation}, (3) Entropy-Aware On-Policy Distillation (EOPD)~\citep{jin2026entropy}, an OPD variant that adds a forward KL in high teacher-entropy states to preserve diversity and teacher distributional support, (4) Skew-KL~\citep{kodistillm}, equivalent to ConMix-FKL with a constant interpolation coefficient of $\alpha_t= 0.1$, and (5) TAID~\citep{shing2025taid}, a method that interpolates student-teacher distributions with a temporally adaptive schedule.

\begin{table*}[h]
\centering
\caption{Offline distillation results on Qwen3-1.7B using MATH. Results are reported as mean$_{\pm \text{std}}$ across three seeds. \textbf{Bold} marks the best result and \underline{underlining} marks the second-best result; ties receive the same formatting.}
\label{tab:qwen3_base_math_sft}
\resizebox{\textwidth}{!}{%
\begin{tabular}{ll cccccc}
\toprule
\multirow{2}{*}{\textbf{Benchmark}} & \multirow{2}{*}{\textbf{Metric}} & \multicolumn{6}{c}{\textbf{Qwen3-1.7B} (MATH)} \\
\cmidrule(lr){3-8}
& & \textbf{\makecell{Student-\\generated}}
  & \textbf{\makecell{Teacher-\\generated}}
  & \textbf{SKD} & \textbf{EAFT}
  & \cellcolor{blue!10}\textbf{ConMix-SFT}
  & \cellcolor{blue!10}\textbf{GeoMix-SFT} \\
\midrule
\multicolumn{8}{l}{\textbf{\textit{Mathematical reasoning}}} \\

\midrule
\multirow{3}{*}{AIME24}
& Avg@64
& \underline{11.06}$_{\pm 0.63}$
& 8.70$_{\pm 0.42}$
& 9.49$_{\pm 0.53}$
& 10.17$_{\pm 0.30}$
& \cellcolor{blue!10}10.57$_{\pm 0.66}$
& \cellcolor{blue!10}\textbf{11.54}$_{\pm 0.55}$ \\
& Pass@8
& 24.97$_{\pm 0.06}$
& 23.96$_{\pm 1.75}$
& \underline{26.47}$_{\pm 1.32}$
& 26.07$_{\pm 2.17}$
& \cellcolor{blue!10}25.35$_{\pm 2.98}$
& \cellcolor{blue!10}\textbf{27.59}$_{\pm 0.27}$ \\
& Pass@64
& 41.67$_{\pm 2.35}$
& 46.67$_{\pm 5.09}$
& \underline{47.78}$_{\pm 5.09}$
& 45.56$_{\pm 5.09}$
& \cellcolor{blue!10}43.34$_{\pm 9.43}$
& \cellcolor{blue!10}\textbf{50.00}$_{\pm 0.00}$ \\

\midrule
\multirow{3}{*}{GSM8K}
& Avg@64  & 80.94$_{\pm 0.10}$ & 79.98$_{\pm 0.20}$ & 79.73$_{\pm 0.23}$ & 79.67$_{\pm 0.16}$ & \cellcolor{blue!10}\textbf{81.12}$_{\pm 0.37}$ & \cellcolor{blue!10}\underline{81.03}$_{\pm 0.16}$ \\
& Pass@8  & 93.12$_{\pm 0.14}$ & \underline{94.45}$_{\pm 0.14}$ & 94.23$_{\pm 0.25}$ & 94.23$_{\pm 0.03}$ & \cellcolor{blue!10}\textbf{94.65}$_{\pm 0.37}$ & \cellcolor{blue!10}94.37$_{\pm 0.00}$ \\
& Pass@64 & 97.07$_{\pm 0.23}$ & \textbf{98.20}$_{\pm 0.26}$ & \underline{98.00}$_{\pm 0.20}$ & 97.93$_{\pm 0.31}$ & \cellcolor{blue!10}\underline{98.00}$_{\pm 0.28}$ & \cellcolor{blue!10}\textbf{98.20}$_{\pm 0.28}$ \\
\midrule
\multirow{3}{*}{MATH500}
& Avg@64  & 70.56$_{\pm 0.09}$ & 69.36$_{\pm 0.13}$ & 69.36$_{\pm 0.12}$ & 69.16$_{\pm 0.13}$ & \cellcolor{blue!10}\textbf{70.99}$_{\pm 0.44}$ & \cellcolor{blue!10}\underline{70.90}$_{\pm 0.49}$ \\
& Pass@8  & 88.76$_{\pm 0.38}$ & 89.45$_{\pm 0.10}$ & \underline{89.55}$_{\pm 0.08}$ & 89.31$_{\pm 0.12}$ & \cellcolor{blue!10}\textbf{90.00}$_{\pm 0.24}$ & \cellcolor{blue!10}89.47$_{\pm 0.18}$ \\
& Pass@64 & 95.33$_{\pm 0.64}$ & 96.00$_{\pm 0.61}$ & 95.47$_{\pm 0.61}$ & \underline{96.13}$_{\pm 0.61}$ & \cellcolor{blue!10}\textbf{96.30}$_{\pm 0.14}$ & \cellcolor{blue!10}95.40$_{\pm 0.28}$ \\
\midrule
\multirow{3}{*}{Minerva}
& Avg@64  & \underline{27.94}$_{\pm 0.10}$ & 27.34$_{\pm 0.26}$ & 27.58$_{\pm 0.37}$ & 27.14$_{\pm 0.14}$ & \cellcolor{blue!10}27.84$_{\pm 0.25}$ & \cellcolor{blue!10}\textbf{27.97}$_{\pm 0.11}$ \\
& Pass@8  & 43.85$_{\pm 0.54}$ & 44.74$_{\pm 0.25}$ & 45.10$_{\pm 0.21}$ & 45.35$_{\pm 0.29}$ & \cellcolor{blue!10}\underline{46.38}$_{\pm 0.02}$ & \cellcolor{blue!10}\textbf{46.52}$_{\pm 0.15}$ \\
& Pass@64 & 54.04$_{\pm 0.97}$ & 54.78$_{\pm 1.14}$ & 56.62$_{\pm 0.98}$ & \underline{56.86}$_{\pm 1.29}$ & \cellcolor{blue!10}\textbf{56.99}$_{\pm 0.00}$ & \cellcolor{blue!10}56.80$_{\pm 0.78}$ \\

\midrule
\multicolumn{8}{l}{\textbf{\textit{General capabilities}}} \\
\midrule
HellaSwag
& Avg@8 & 50.59$_{\pm 0.03}$ & 51.59$_{\pm 0.32}$ & 51.76$_{\pm 0.13}$ & 51.91$_{\pm 0.32}$ & \cellcolor{blue!10}\textbf{55.51}$_{\pm 0.68}$ & \cellcolor{blue!10}\underline{55.07}$_{\pm 0.66}$ \\
IFEval
& Avg@8 & \underline{41.71}$_{\pm 0.65}$ & 38.82$_{\pm 1.19}$ & 38.32$_{\pm 1.80}$ & 33.76$_{\pm 0.83}$ & \cellcolor{blue!10}41.50$_{\pm 3.88}$ & \cellcolor{blue!10}\textbf{41.96}$_{\pm 1.57}$ \\
MMLU
& Avg@8
& \textbf{63.30}$_{\pm 0.20}$
& 61.00$_{\pm 0.97}$
& 61.40$_{\pm 0.70}$
& 60.80$_{\pm 0.50}$
& \cellcolor{blue!10}62.20$_{\pm 0.50}$
& \cellcolor{blue!10}\underline{62.70}$_{\pm 0.40}$ \\
\bottomrule
\end{tabular}%
}
\end{table*}

\section{Evaluation Results} \label{sec:math_performance}



\begin{table*}[t]
\centering
\caption{Online distillation results on Qwen3-1.7B-Base using MATH. Results are reported as mean$_{\pm \text{std}}$ across three seeds;  formatting follows Table~\ref{tab:qwen3_base_math_sft}.}
\label{tab:qwen3_base_math_kd}
\resizebox{\textwidth}{!}{%
\begin{tabular}{ll ccccccc}
\toprule
\multirow{2}{*}{\textbf{Benchmark}}
& \multirow{2}{*}{\textbf{Metric}}
& \multicolumn{7}{c}{\textbf{Qwen3-1.7B-Base} (MATH)} \\
\cmidrule(lr){3-9}
&
& \textbf{FKL}
& \textbf{OPD}
& \textbf{EOPD}
& \textbf{Skew-KL}
& \textbf{TAID}
& \cellcolor{blue!10}\textbf{ConMix-FKL}
& \cellcolor{blue!10}\textbf{GeoMix-FKL} \\
\midrule
\multicolumn{9}{l}{\textbf{\textit{Mathematical reasoning}}} \\

\midrule
\multirow{3}{*}{AIME24}
& Avg@64
& 5.28$_{\pm 0.44}$
& \underline{7.63}$_{\pm 0.26}$
& 7.20$_{\pm 0.43}$
& 4.91$_{\pm 0.11}$
& 5.89$_{\pm 0.42}$
& \cellcolor{blue!10}6.75$_{\pm 0.59}$
& \cellcolor{blue!10}\textbf{8.15}$_{\pm 0.27}$ \\
& Pass@8
& 18.17$_{\pm 0.73}$
& 21.15$_{\pm 0.85}$
& 19.96$_{\pm 1.02}$
& 17.25$_{\pm 0.62}$
& 18.41$_{\pm 2.68}$
& \cellcolor{blue!10}\underline{21.37}$_{\pm 0.38}$
& \cellcolor{blue!10}\textbf{22.67}$_{\pm 0.95}$ \\
& Pass@64
& 35.56$_{\pm 1.93}$
& 35.57$_{\pm 1.77}$
& 34.44$_{\pm 1.93}$
& \underline{36.67}$_{\pm 6.67}$
& \underline{36.67}$_{\pm 3.34}$
& \cellcolor{blue!10}35.55$_{\pm 3.85}$
& \cellcolor{blue!10}\textbf{39.72}$_{\pm 2.87}$ \\

\midrule
\multirow{3}{*}{GSM8K}
& Avg@64
& 79.22$_{\pm 0.12}$
& \underline{81.57}$_{\pm 0.10}$
& 81.55$_{\pm 0.08}$
& 79.10$_{\pm 0.05}$
& 80.02$_{\pm 0.14}$
& \cellcolor{blue!10}80.36$_{\pm 0.04}$
& \cellcolor{blue!10}\textbf{81.62}$_{\pm 0.04}$ \\
& Pass@8
& 95.48$_{\pm 0.22}$
& 95.37$_{\pm 0.08}$
& \textbf{95.72}$_{\pm 0.17}$
& 95.42$_{\pm 0.04}$
& 95.58$_{\pm 0.15}$
& \cellcolor{blue!10}95.54$_{\pm 0.07}$
& \cellcolor{blue!10}\underline{95.59}$_{\pm 0.15}$ \\
& Pass@64
& \textbf{99.47}$_{\pm 0.12}$
& 99.12$_{\pm 0.15}$
& 99.13$_{\pm 0.12}$
& 99.33$_{\pm 0.23}$
& 99.07$_{\pm 0.42}$
& \cellcolor{blue!10}99.27$_{\pm 0.31}$
& \cellcolor{blue!10}\underline{99.41}$_{\pm 0.08}$ \\

\midrule
\multirow{3}{*}{MATH500}
& Avg@64
& 59.68$_{\pm 0.24}$
& \underline{64.43}$_{\pm 0.11}$
& 63.06$_{\pm 0.07}$
& 57.31$_{\pm 0.16}$
& 60.47$_{\pm 0.07}$
& \cellcolor{blue!10}62.13$_{\pm 0.07}$
& \cellcolor{blue!10}\textbf{64.45}$_{\pm 0.15}$ \\
& Pass@8
& 84.65$_{\pm 0.39}$
& \underline{85.98}$_{\pm 0.04}$
& 85.40$_{\pm 0.05}$
& 83.61$_{\pm 0.04}$
& 84.78$_{\pm 0.25}$
& \cellcolor{blue!10}85.66$_{\pm 0.07}$
& \cellcolor{blue!10}\textbf{86.04}$_{\pm 0.08}$ \\
& Pass@64
& \underline{94.13}$_{\pm 0.31}$
& 93.79$_{\pm 0.14}$
& 93.47$_{\pm 0.61}$
& 93.40$_{\pm 1.06}$
& \textbf{94.27}$_{\pm 0.23}$
& \cellcolor{blue!10}94.07$_{\pm 1.29}$
& \cellcolor{blue!10}93.95$_{\pm 0.22}$ \\

\midrule
\multirow{3}{*}{Minerva}
& Avg@64
& 23.68$_{\pm 0.20}$
& \textbf{24.81}$_{\pm 0.04}$
& 24.71$_{\pm 0.13}$
& 22.65$_{\pm 0.34}$
& 23.48$_{\pm 0.23}$
& \cellcolor{blue!10}24.29$_{\pm 0.38}$
& \cellcolor{blue!10}\underline{24.79}$_{\pm 0.11}$ \\
& Pass@8
& \underline{44.90}$_{\pm 0.20}$
& 44.23$_{\pm 0.11}$
& 44.84$_{\pm 0.44}$
& 44.09$_{\pm 0.26}$
& 43.96$_{\pm 0.17}$
& \cellcolor{blue!10}\underline{44.90}$_{\pm 0.69}$
& \cellcolor{blue!10}\textbf{45.17}$_{\pm 0.28}$ \\
& Pass@64
& \textbf{58.21}$_{\pm 1.18}$
& 55.96$_{\pm 0.28}$
& 57.47$_{\pm 1.29}$
& 57.35$_{\pm 1.11}$
& 57.11$_{\pm 0.76}$
& \cellcolor{blue!10}57.60$_{\pm 1.18}$
& \cellcolor{blue!10}\underline{58.05}$_{\pm 0.23}$ \\

\midrule
\multicolumn{9}{l}{\textbf{\textit{General capabilities}}} \\
\midrule

HellaSwag
& Avg@8
& 61.33$_{\pm 0.07}$
& 61.18$_{\pm 0.03}$
& \underline{61.39}$_{\pm 0.04}$
& 61.10$_{\pm 0.07}$
& 61.28$_{\pm 0.06}$
& \cellcolor{blue!10}61.34$_{\pm 0.03}$
& \cellcolor{blue!10}\textbf{61.42}$_{\pm 0.02}$ \\

IFEval
& Avg@8
& 23.48$_{\pm 0.49}$
& 24.21$_{\pm 1.12}$
& 24.46$_{\pm 1.57}$
& 23.41$_{\pm 1.20}$
& 24.28$_{\pm 1.76}$
& \cellcolor{blue!10}\underline{24.95}$_{\pm 0.74}$
& \cellcolor{blue!10}\textbf{25.26}$_{\pm 0.77}$ \\

MMLU
& Avg@8
& 47.90$_{\pm 0.40}$
& 47.80$_{\pm 1.10}$
& 48.20$_{\pm 0.40}$
& 46.50$_{\pm 0.30}$
& 48.00$_{\pm 0.20}$
& \cellcolor{blue!10}\underline{48.90}$_{\pm 0.40}$
& \cellcolor{blue!10}\textbf{49.30}$_{\pm 0.20}$ \\

\bottomrule
\end{tabular}%
}
\end{table*}

\subsection{Main results}
\paragraph{Offline distillation.}
As shown in Table~\ref{tab:qwen3_base_math_sft}, student-generated traces often improve per-sample accuracy by aligning with the student's inference distribution, whereas we find that teacher-derived supervision can improve Pass@64 by exposing the student to otherwise underexplored reasoning trajectories. In contrast, GeoMix-SFT combines both behaviours, improving +1.05 Avg@64 over teacher-based baselines (i.e., teacher-generated, SKD, and EAFT) and +1.22 Pass@8 and +2.76 Pass@64 over the student-generated baseline, averaged across six mathematical benchmarks (detailed in Table~\ref{tab:qwen3_17b_math_sft}).
HellaSwag, IFEval, and MMLU reveal clear differences in how well each supervision strategy preserves general capabilities. Teacher-derived traces degrade IFEval performance,
while entropy-aware mixtures consistently rank among the strongest methods, likely due to their more student-aligned trajectories.

\paragraph{Online distillation.}
As reported in Table~\ref{tab:qwen3_base_math_kd}, GeoMix-FKL achieves the strongest performance improvement and performs comparably to OPD, likely due to their similar training dynamics (described in Section~\ref{sec:online_training_dynamics}). A more detailed comparison of aggregate performance across the six mathematical benchmarks is provided in Table~\ref{tab:qwen3_17b_base_math_kd}. More generally, ConMix-FKL and GeoMix-FKL consistently outperform standard FKL, showing that entropy-aware target shaping improves upon the objective that our method builds on. Finally, ConMix-FKL also consistently outperforms Skew-KL, suggesting an advantage for entropy-adaptive over fixed interpolation.
Regarding general capabilities, GeoMix-FKL achieves the strongest performance in the three benchmarks. Among the baselines, EOPD and TAID best preserve broader capabilities; nevertheless, GeoMix-FKL achieves stronger mathematical reasoning performance.

\subsection{Ablation studies}
We summarise the main ablation results here, with all results reported in Appendix~\ref{app:qwen3_app_results}-\ref{app:extra_models}.

\paragraph{Smaller student.}
With Qwen3-0.6B non-thinking mode, we observe the same overall trends as for Qwen3-1.7B. Tables~\ref{tab:qwen3_06b_math_sft} and~\ref{tab:qwen3_06b_base_math_kd} show that entropy-aware mixing improves mathematical reasoning in both offline and online distillation. Relative to the evaluated baselines, GeoMix-SFT and GeoMix-FKL achieve the highest Avg@64, while their Pass@64 performance remains comparable to that of other teacher baselines. The method therefore remains effective under a greater student-teacher capacity gap.

\paragraph{Code generation.}
To test whether the gains extend beyond mathematical reasoning, we evaluate Qwen3-1.7B non-thinking on LeetCodeDataset (Tables~\ref{tab:qwen3_code} and~\ref{tab:qwen3_base_code}). In the offline setting, both ConMix-SFT and GeoMix-SFT outperform the teacher baselines (i.e., teacher-generated, SKD, and EAFT) on Avg@64 and Pass@1, while remaining comparable on Pass@64. Here, Avg@64 denotes the mean pass rate across 64 samples, whereas Pass@$k$ measures whether at least one of $k$ responses passes all unit tests. In the online setting, GeoMix-FKL achieves the highest mean performance across all metrics, with TAID producing comparable results. The same benefits of entropy-aware mixing therefore carry over to code generation.

\paragraph{Transfer across model families.}
We also apply the online distillation setup to LFM2.5-350M with MATH and Granite-3.2-2B-Instruct with DAPO-Math-14k (Appendix~\ref{app:extra_models}). Entropy-aware mixing remains competitive with the strongest baselines in both settings. On LFM2.5-350M, GeoMix-FKL improves over the best baseline by $+1.06$ Avg@64 and $+1.91$ Pass@8. On Granite-3.2-2B-Instruct, smaller improvements of $+0.15$ Avg@64 and $+0.03$ Pass@8 are observed. The gains are larger for the smaller students (LFM2.5-350M and Qwen3-0.6B), for which the student-teacher capacity gap is greater.

\paragraph{Alternative mixture distributions.}
We examine whether other mixture forms defined by AMiD~\citep{shin2026amid} benefit from entropy-aware interpolation. For $\beta \in \{-5, -3, 0\}$, replacing the fixed interpolation coefficient $\alpha_t=0.1$ with an entropy-aware coefficient improves performance across several metrics (Table~\ref{tab:full_qwen3_base_amid}). At $\beta=0$, Avg@64 increases from $34.04\pm0.04$ to $35.95\pm0.05$. We therefore conclude that our results are not limited to the convex and geometric mixtures considered in the main experiments.

\paragraph{Alternative distillation objective.}
Finally, we briefly test the effect of replacing the forward-KL objective with the reverse-KL objective from OPD~\citep{lu2025onpolicydistillation}, to assess whether adaptive mixtures also provide benefits under other divergences. The results in Appendix~\ref{app:opd} show that GeoMix-OPD remains competitive with standard OPD while improving aggregate Pass@64.

\section{Understanding Entropy-Aware Mixing}
We further analyse the resulting offline dataset, the dynamics of online distillation, and alternative entropy-based interpolation strategies for our approach.


\subsection{Student-Aligned Supervision} \label{sec:student_aligned_supervision}
\begin{figure}[tb]
    \centering
    \begin{subfigure}[t]{0.34\textwidth}
        \centering
        \includegraphics[width=\textwidth, valign=t]{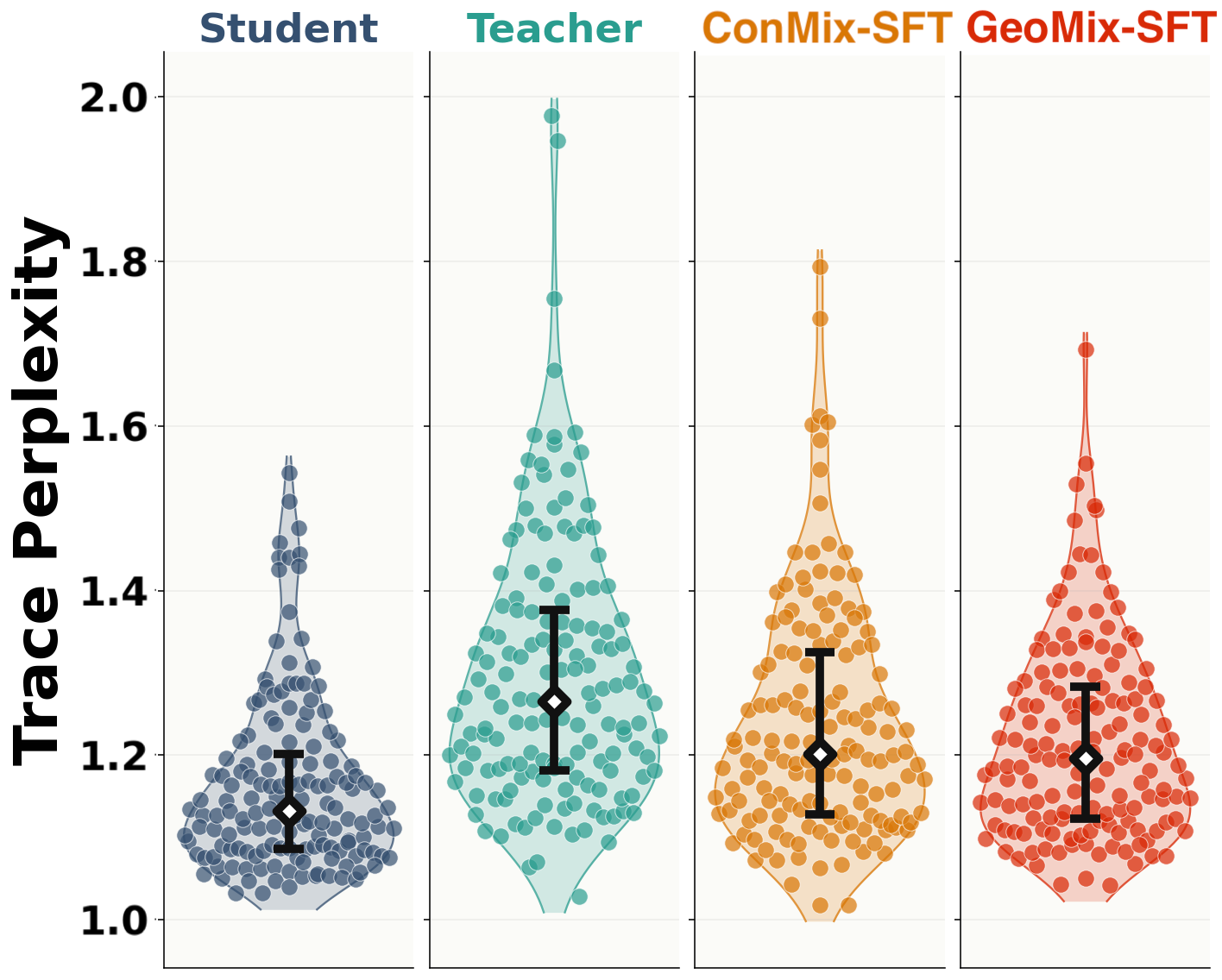}
        \caption{Student trace perplexity under each target distribution (student, teacher, and entropy-aware mixtures). Lower values indicate higher confidence.}
        \label{fig:trace_per}
    \end{subfigure}
    \hfill
    \begin{subfigure}[t]{0.64\textwidth}
        \centering
        \includegraphics[width=\textwidth, valign=t]{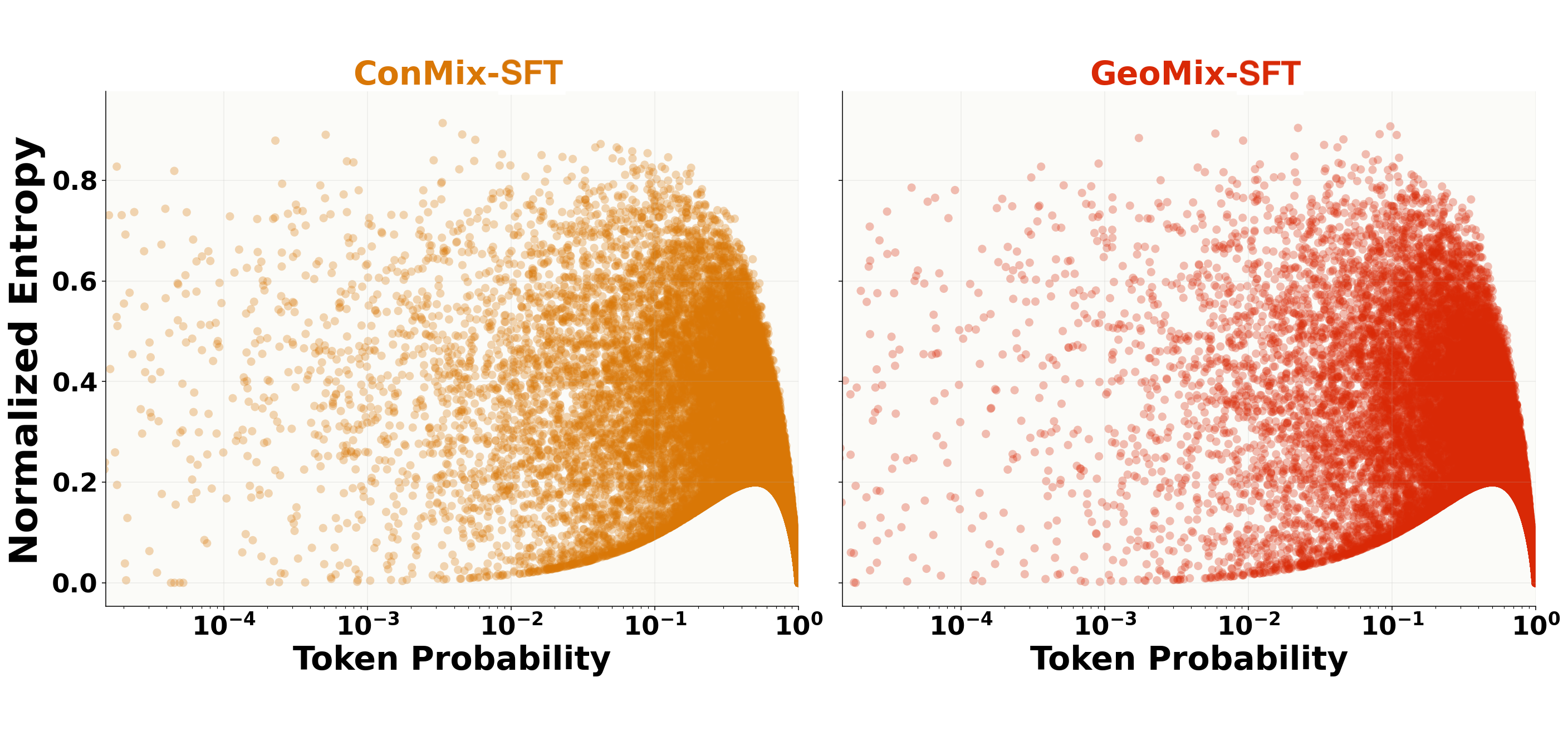}
        \caption{Normalised student entropy vs token probability for ConMix-SFT (left) and GeoMix-SFT (right) traces.}
        \label{fig:entropy_prob_2}
    \end{subfigure}
    \caption{Token-level analysis of entropy-aware traces (same setup as Figure~\ref{fig:confident_conflicts}). Entropy-aware speculative decoding produces traces that are closer to the student distribution while low-probability tokens now occur in higher-entropy states.}
    \label{fig:confident_conflicts_2}
\end{figure}
\paragraph{How Entropy-Aware Mixing changes the supervision.}
As shown in Figure~\ref{fig:trace_per}, teacher-generated traces have substantially higher perplexity under the student model than student-generated traces, indicating a larger distributional mismatch. In contrast, traces sampled from ConMix-SFT and GeoMix-SFT exhibits lower perplexity and lie between the student and teacher distributions. ConMix-SFT preserves more teacher-like characteristics due to its linear interpolation of distributions, whereas GeoMix-SFT penalises large student-teacher disagreements, producing traces that are closer to the student's distribution.

\paragraph{Why this improves the optimisation stability.}
Figure~\ref{fig:entropy_prob_2} shows that entropy-aware mixing largely removes the tail of low-probability tokens in low-entropy student states observed with teacher supervision (Figure~\ref{fig:entropy_prob}). Instead, low-probability tokens occur primarily when the student is uncertain, concentrating larger distributional shifts in states where its predictions are less established. This also has an optimisation interpretation: FKL updates can contain large density-ratio coefficients when the student assigns very low probability to teacher-supported tokens~\citep{kodistillm}. For the generalised AMiD target, we show in Appendix~\ref{app:amid_gradient} that entropy-aware interpolation reduces the contribution of large probability mismatches by using smaller teacher mixing weights in low-entropy states while retaining stronger teacher guidance as student entropy increases. These results indicate greater optimisation stability of our entropy-aware mixture compared with standard FKL.

\paragraph{Confident student errors.} 
Low-entropy regimes do not necessarily imply correctness, but rather confidence. Confident conflicts, as discussed in Section~\ref{sec:confident_conflicts}, typically reflect disagreements with the teacher that are not limited to final answers. Our analysis in Appendix~\ref{app:confident_conflicts} shows that such disagreements do not constitute a systematic failure mode of our entropy-based interpolation. On prompts and prefixes associated with incorrect low-entropy student trajectories, entropy-aware mixtures substantially outperform student generations and remain close to teacher performance, indicating that they can recover even in the presence of confident student errors.

\subsection{Online Training Dynamics} \label{sec:online_training_dynamics}
\begin{figure}
    \centering
    \includegraphics[width=1\linewidth]{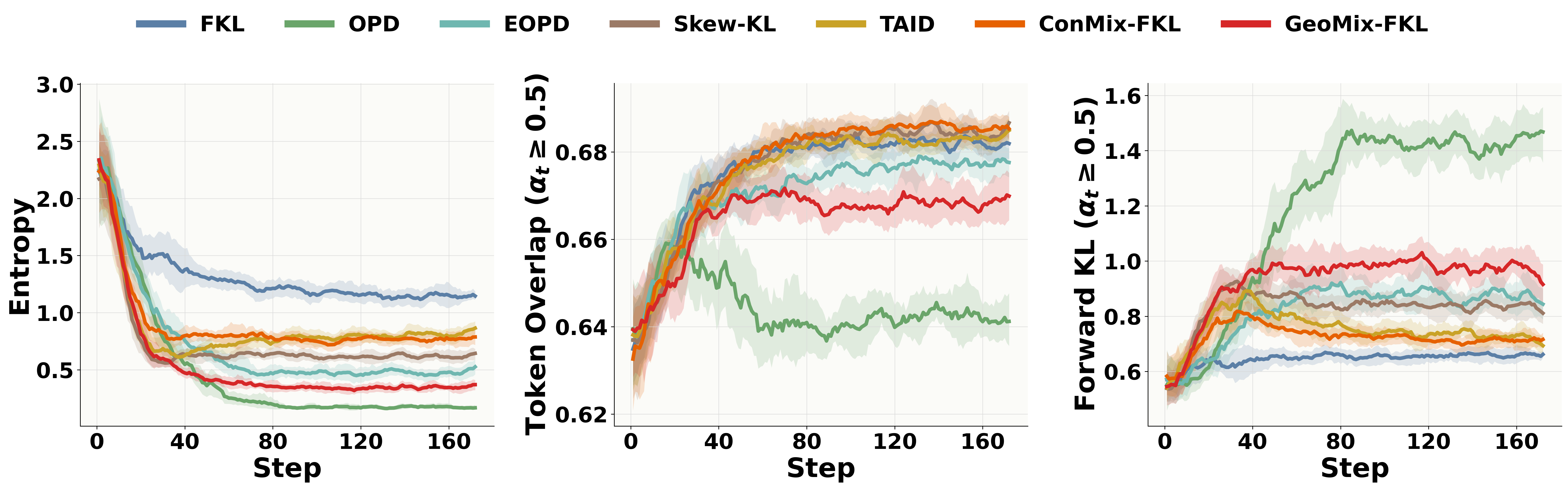}
    \caption{Online distillation training dynamics from Qwen3-1.7B-Base. \textbf{Left}: policy entropy over training. \textbf{Middle}: student-teacher top-64 token overlap on high-\(\alpha_t\) states (\(\alpha_t \geq 0.5\)). \textbf{Right}: forward KL against the teacher on high-entropy states. Shaded regions denote standard deviation across seeds.}
    \label{fig:online_token_analysis}
\end{figure}

Figure~\ref{fig:online_token_analysis} shows how the different online distillation objectives shape the student distribution during training. Standard FKL maintains the highest policy entropy and remains closest to the teacher, with high student-teacher token overlap and low forward KL on high-\(\alpha_t\) states. In contrast, OPD produces the sharpest policy, together with the lowest token overlap and the largest forward KL in high-\(\alpha_t\) states. Skew-KL, TAID, ConMix-FKL, and EOPD fall between these two regimes, reducing entropy relative to FKL while retaining substantially more teacher agreement than OPD.

GeoMix-FKL exhibits a different behaviour: its entropy dynamics approach the lower-entropy regime of OPD, while its student-teacher token overlap remains substantially higher. Its forward KL on high-\(\alpha_t\) states is also higher than for the other FKL-based methods, which is consistent with the geometric mixture concentrating probability on overlapping tokens between the student and teacher rather than matching the full teacher distribution directly. GeoMix-FKL changes the composition of the target rather than merely weakening teacher supervision, and its downstream performance suggests better generalisation.

\subsection{Effect of the Entropy-Based Interpolation Schedule}
\label{sec:entropy_transform}
\begin{figure}[tb]
    \centering
    \begin{subfigure}[t]{0.47\textwidth}
        \centering
        \includegraphics[width=\textwidth, valign=t]{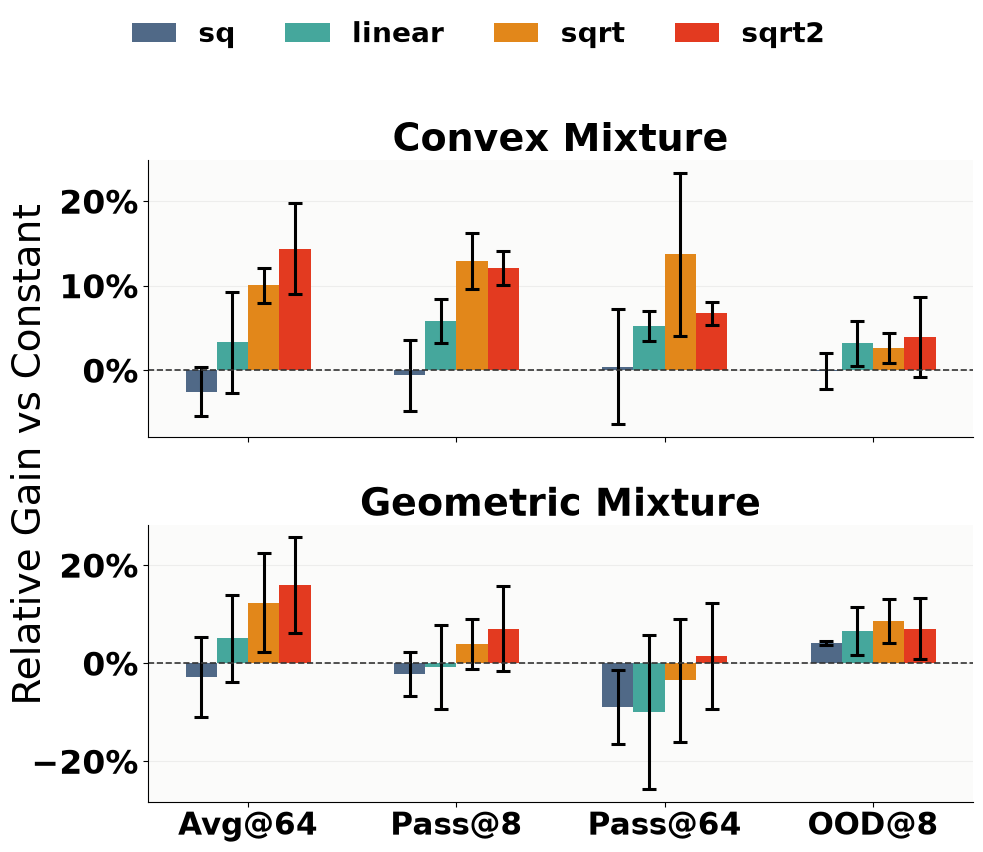}
        \subcaption{Offline distillation on Qwen3-1.7B.}
        \label{fig:alpha_offline}
    \end{subfigure}
    \hfill
    \begin{subfigure}[t]{0.47\textwidth}
        \centering
        \includegraphics[width=\textwidth, valign=t]{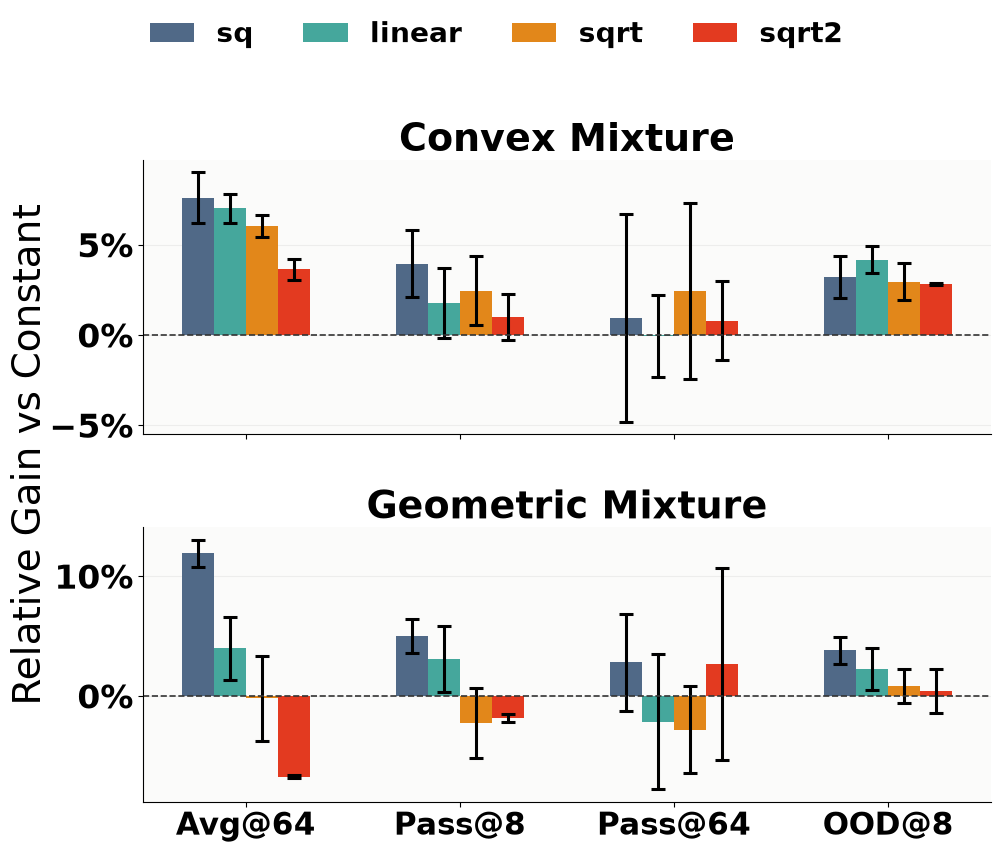}
        \subcaption{Online distillation on Qwen3-1.7B-Base.}
        \label{fig:alpha_online}
    \end{subfigure}
    \caption{Relative performance change under different entropy transformations \(f\) relative to a constant coefficient; error bars denote standard deviation across three seeds.}
    \label{fig:alpha}
\end{figure}

Figure~\ref{fig:alpha} studies how the mapping from student entropy to the interpolation coefficient $\alpha_t$ affects performance. We compare a fixed coefficient $f(x)=0.1$, a convex schedule $f(x)=x^2$ (\texttt{sq}), a linear schedule $f(x)=x$ (\texttt{linear}), and two concave schedules, $f(x)=x^{1/2}$ (\texttt{sqrt}) and $f(x)=x^{1/4}$ (\texttt{sqrt2}), while keeping the remaining training setup fixed. This ablation tests both whether token-wise adaptation matters and where teacher guidance should be concentrated across different levels of student entropy.

\textbf{Offline distillation.}
Figure~\ref{fig:alpha_offline} shows that offline trace generation favours concave schedules. In both mixture forms, \texttt{sqrt} and \texttt{sqrt2} consistently outperform the linear schedule, while the convex schedule performs worse. Concave mappings place more teacher weight on lower-entropy states than the linear schedule, while still keeping teacher influence small when the student is highly confident. Notably, the fixed coefficient also underperforms, despite approximately matching the average teacher weight of the linear schedule ($\mathbb{E}[\alpha_t]\approx0.1$). Thus, performance depends not only on the overall amount of teacher influence, but also on \emph{where} that influence is applied across tokens.

\textbf{Online distillation.}
Figure~\ref{fig:alpha_online} shows the opposite pattern. Here, the linear and convex schedules perform best, whereas the concave schedules generally have degraded performance. This suggests that on-policy distillation benefits from concentrating teacher guidance in higher-entropy states rather than introducing it more strongly at lower levels. Again, the fixed-$\alpha$ baseline underperforms the strongest adaptive schedules.

Therefore, the preferred schedule depends on the training regime: offline trace generation benefits from earlier teacher influence as uncertainty increases, whereas online distillation benefits from concentrating teacher influence in the most uncertain states.

\section{Related Work}
\paragraph{Forgetting and distributional mismatch in LLM post-training.}
Catastrophic forgetting has traditionally been mitigated through parameter constraints, replay, or regularisation~\citep{Kirkpatrick_2017,li2017learningforgetting,lopezpaz2022gradientepisodicmemorycontinual}. In LLMs, similar issues arise during continual pre-training, instruction tuning, alignment, and distillation, where increasing specialisation can degrade general capabilities~\citep{betley2026training,luo2025empiricalstudycatastrophicforgetting,ouyang2022traininglanguagemodelsfollow,shi2024continuallearninglargelanguage}. Recent work attributes this degradation to off-policy supervision: training on demonstrations from a different distribution may reshape the student's representations away from its output distribution \citep{agarwal2024policy,li2026rethinkingonpolicydistillationlarge}. On-policy methods narrow this gap by training on student-generated trajectories under teacher guidance~\citep{agarwal2024policy,chen2025retainingdoingroleonpolicy,minillm,lu2025onpolicydistillation}; similarly self-distillation methods instead use feedback-conditioned student variants as in-context teachers \citep{hubotter2026reinforcementlearningselfdistillation,yang2024selfdistillationbridgesdistributiongap}. Other forms of mixed supervision, such as interleaving SFT and RL, requires careful baseline calibration to outperform sequential SFT-then-RL~\citep{limozin2026sftthenrl}. These findings motivate more efficient teacher-guided learning that improves generalisation.

\paragraph{Entropy-guided supervision.}
Recent analyses link general-capability degradation to high loss on low-probability tokens, motivating SFT objectives that downweight their influence~\citep{lin2026sftdoesnthurtgeneral,wu2026generalisationsftreinforcementlearning,zhang2026onpolicysftdistributiondiscriminant}. DFT~\citep{wu2026generalisationsftreinforcementlearning} and TALR~\citep{lin2026sftdoesnthurtgeneral} reweight tokens based on token probability, whereas EAFT~\citep{diao2026entropy} uses entropy to suppress updates in low-entropy states. Aligned with this view, \citet{wang20258020rulehighentropyminority} show that RLVR gains are driven by a small subset of high-entropy tokens, while low-entropy tokens contribute little or can even hurt performance. Overall, these works suggest that token-level uncertainty can provide guidance where learning should be concentrated~\citep{TokUR}. In our work, rather than reweighting the teacher response, we use entropy to construct an adaptive student-teacher distribution. Entropy therefore determines not only how strongly the student learns, but also \emph{what} it imitates at each token.

\paragraph{Student-aligned generation.}
SKD~\citep{xu2024speculative} interleaves student proposals with teacher corrections, producing samples closer to the student's inference-time distribution than fixed teacher traces. RSD~\citep{kim2025wordsreasoningtracestailored} reverses these roles: the teacher proposes tokens while the student filters low-probability continuations, yielding lower-surprisal, student-friendly traces. EASD~\citep{su2025entropyawarespeculativedecodingimproved} uses entropy and top-$K$ overlap to resample when both models are uncertain and their candidate sets overlap. Unlike these rule-based trace-generation procedures, our entropy-conditioned student-teacher target distribution preserves confident student behaviour while injecting teacher guidance in uncertain states.

\section{Conclusion} \label{sec:conclusion}
We introduced Entropy-Aware Mixing, a token-level distillation strategy that uses student entropy to interpolate between student and teacher distributions. Rather than distilling directly from the teacher, the mixed distribution stays close to the student in low-entropy states and shifts toward the teacher as student entropy increases. We test our method on: (i) supervised fine-tuning with static traces generated from the mixture, and (ii) on-policy forward-KL where the mixtures acts as the target. Across both settings, our methods outperform traditional fixed-teacher supervision in both mathematical reasoning and general capabilities. Further analysis reveals that the optimal interpolation depends on the training regime: offline training benefits from earlier teacher influence, whereas online distillation favours concentrating it in high-entropy states. 

\paragraph{Limitations and open questions.}
It remains unclear which tokens most strongly reshape the student's shared representation; predictive entropy may simply serve as a proxy for these influential updates rather than their underlying cause. Developing principled or learned interpolation schedules is therefore an important direction, particularly given the different entropy-based transformations used in offline and online training.  Although we observed that preserving confident behaviours does not necessarily lead to worse performance, future work could combine verifiers to better determine when teacher supervision should override the student (especially in safety-critical domains). Our evaluation was also limited to student-teacher pairs from the same model family; extending such analysis to models from different families could help demonstrate the effectiveness of preserving student behaviour.

\bibliographystyle{plainnat}
\bibliography{neurips_2026}

\clearpage
\appendix
\startcontents[appendices]
\section*{Appendix Overview}
\printcontents[appendices]{}{1}{}
\clearpage

\newpage
\section{Background: Supervised Fine-Tuning and Distillation Objectives} \label{app:background}
\subsection{Supervised Fine-Tuning}
Let $\mathcal{D} = \{(\mathbf{x}, \mathbf{y})_i\}_{i=1}^N$ denote an offline dataset of input--output pairs, where each target sequence $\mathbf{y}_i$ is obtained either from human annotations or from off-policy samples generated by a stronger model. For simplicity, we omit the index $i$ and write $(\mathbf{x}, \mathbf{y}) \sim \mathcal{D}$, where $\mathbf{y} = (y_1, \dots, y_T)$ denotes a target sequence of length $T = |\mathbf{y}|$.

At decoding step $t$, given context $\mathbf{c}_t = (\mathbf{x}, \mathbf{y}_{<t})$, the standard token-level cross-entropy objective minimises the forward KL divergence between the one-hot target distribution $q_t(v) = \mathbb{I}[v = y_t]$ and the model distribution $\pi_\theta(\cdot \mid \mathbf{c}_t)$. The per-sequence loss is
\begin{equation}
\mathcal{L}^{\mathrm{CE}}(\mathbf{x}, \mathbf{y}; \theta)
= - \frac{1}{T} \sum_{t=1}^{T} \log \pi_\theta(y_t \mid \mathbf{c}_t)
= \frac{1}{T} \sum_{t=1}^{T} D_{\mathrm{KL}}\!\left(q_t \,\|\, \pi_\theta(\cdot \mid \mathbf{c}_t)\right).
\end{equation}
Thus, low-probability tokens induce large updates, which can be disruptive based on the model's entropy \citep{diao2026entropy, wu2026generalisationsftreinforcementlearning}.

\subsection{Knowledge Distillation}
In contrast, distillation transfers token-level behaviours using probability distributions typically produced by a larger teacher model \(\pi_{\mathrm{te}}\) \citep{hinton2015distillingknowledgeneuralnetwork}.

\paragraph{Off-Policy Objectives.}
For a given target sequence $\mathbf{y}$ from a sample $(\mathbf{x}, \mathbf{y}) \in \mathcal{D}$, we compute the token-level Kullback-Leibler (KL) divergence between the teacher and student distributions over the vocabulary as:
\begin{equation} \label{eq:forward_kl}
    D_{\mathrm{KL}}\!\left(\pi_{\mathrm{te}}(\cdot \mid \mathbf{c}_t) \,\|\, \pi_\theta(\cdot \mid \mathbf{c}_t)\right)
    = \mathbb{E}_{y' \sim \pi_{\mathrm{te}}(\cdot \mid \mathbf{c}_t)}
    \left[
    \log \frac{\pi_{\mathrm{te}}(y' \mid \mathbf{c}_t)}{\pi_\theta(y' \mid \mathbf{c}_t)}
    \right].
\end{equation}

Since the KL divergence is asymmetric, it yields two distinct optimisation objectives with fundamentally different inductive biases~\citep{minka2005divergence}:
\begin{enumerate}
    \item \textbf{Forward KL (Teacher-to-Student):} Minimising Equation \eqref{eq:forward_kl} over all target tokens in the dataset forces the student to assign probability mass wherever the teacher's probability is non-zero, resulting in \emph{mode-covering} behaviour.
    
    \item \textbf{Reverse KL (Student-to-Teacher):}  
    Conversely, minimising $D_{\mathrm{KL}}\!\left(\pi_{\theta}(\cdot \mid \mathbf{c}_t) \,\|\, \pi_{\mathrm{te}}(\cdot \mid \mathbf{c}_t)\right)$ across all target tokens encourages \emph{mode-seeking} behaviour, as it heavily penalises the student for allocating mass where the teacher's probability is low.
\end{enumerate}

\paragraph{On-Policy Distillation.}
Recent distillation methods use on-policy sampling to train on the contexts naturally induced by the student policy. Let \(\mathbf{x}\sim\mathcal{D}\) be a prompt, \(\mathbf{y}\sim\pi_\theta(\cdot\mid\mathbf{x})\) a student rollout, and \(\mathbf{c}_t=(\mathbf{x},\mathbf{y}_{<t})\) the corresponding on-policy context. Distillation can then be applied with either KL direction.

\begin{enumerate}
    \item \textbf{Forward KL (Teacher-to-Student):} The on-policy forward-KL objective matches the student to the teacher on student-generated contexts:
\begin{equation}
\label{eq:on_policy_fkl}
    \mathcal{L}_{\mathrm{FKL}}(\theta)
    =
    \mathbb{E}_{\mathbf{x}\sim\mathcal{D},\,
    \mathbf{y}\sim\pi_\theta(\cdot\mid\mathbf{x})}
    \left[
        \frac{1}{T}\sum_{t=1}^{T}
        D_{\mathrm{KL}}\!\left(
            \pi_{\mathrm{te}}(\cdot\mid\mathbf{c}_t)
            \,\|\, 
            \pi_\theta(\cdot\mid\mathbf{c}_t)
        \right)
    \right].
\end{equation}

    \item \textbf{Reverse KL (Student-to-Teacher):}  
    On-Policy Distillation (OPD) \citep{agarwal2024policy, minillm, lu2025onpolicydistillation} is formulated with the reverse KL:
\begin{equation}
\label{eq:opd_obj}
    \mathcal{L}_{\mathrm{OPD}}(\theta)
    =
    \mathbb{E}_{\mathbf{x}\sim\mathcal{D},\,
    \mathbf{y}\sim\pi_\theta(\cdot\mid\mathbf{x})}
    \left[
        \frac{1}{T}\sum_{t=1}^{T}
        D_{\mathrm{KL}}\!\left(
            \pi_\theta(\cdot\mid\mathbf{c}_t)
            \,\|\, 
            \pi_{\mathrm{te}}(\cdot\mid\mathbf{c}_t)
        \right)
    \right].
\end{equation}
For a sampled token \(y_t\sim\pi_\theta(\cdot\mid\mathbf{c}_t)\), we define the
token-level log-ratio $A_t = \log \pi_{\mathrm{te}}(y_t\mid\mathbf{c}_t) - \log \pi_\theta(y_t\mid\mathbf{c}_t)$, which measures how much more the teacher prefers the sampled token than the current student policy. Since
\begin{equation}
    \mathbb{E}_{y_t\sim\pi_\theta(\cdot\mid\mathbf{c}_t)}[A_t]
    =
    -D_{\mathrm{KL}}\!\left(
        \pi_\theta(\cdot\mid\mathbf{c}_t)
        \,\|\, 
        \pi_{\mathrm{te}}(\cdot\mid\mathbf{c}_t)
    \right),
\end{equation}
minimising the reverse KL is equivalent to maximising expected teacher preference along student rollouts:
\begin{equation}
    \max_{\theta}\;
    \mathbb{E}_{\mathbf{x}\sim\mathcal{D},\,
    \mathbf{y}\sim\pi_\theta(\cdot\mid\mathbf{x})}
    \left[
        \frac{1}{T}\sum_{t=1}^{T} A_t
    \right].
\end{equation}
In practice, OPD-style methods often optimise this objective with importance-weighted policy-gradient updates and clipping to avoid overly large policy changes~\citep{lu2025onpolicydistillation}.
\end{enumerate}

\newpage
\section{Background: AMiD} \label{app:amid}
\citet{shin2026amid} formulate knowledge distillation through a generalised family of distributions that interpolate between the student policy $\pi_\theta$ and the teacher policy $\pi_{\mathrm{te}}$. The resulting distribution is referred to as an \emph{assistant distribution}. The framework separates two design choices: (i) a fixed interpolation coefficient $\alpha \in [0,1]$, which determines the relative contributions of the student and teacher and remains constant throughout training, and (ii) a geometry parameter $\beta$, which determines the interpolation geometry between the two distributions.

\paragraph{Interpolation Coefficient.}
Existing mixture-based distillation methods typically use a fixed interpolation coefficient. \citet{kodistillm} set the teacher weight to $0.1$, selected empirically to stabilise the Skew-KL objective by keeping the mixture close to the student while avoiding unstable density ratios. AMiD~\citep{shin2026amid} follows this setting when evaluating its mixture family. In contrast, we study dynamic token-level interpolation coefficients that adapt the induced target distribution to the student state, opening a broader direction for designing context-dependent mixture distributions.

\paragraph{$\beta$-Mixture Distribution.}
At decoding step $t$, let $\mathbf{c}_t=(\mathbf{x},\mathbf{y}_{<t})$ denote the context, where $\mathbf{x}$ is the input prompt and $\mathbf{y}_{<t}$ is the previously generated token sequence. For the next token $y_t$, the unnormalised AMiD distribution is defined as:
\begin{equation} \label{eq:amid}
\widetilde{\pi}_{\mathrm{AMiD}}^{(\beta,\alpha)}
(y_t \mid \mathbf{c}_t)
=
\begin{cases}
\left[
(1-\alpha)
\pi_\theta(y_t \mid \mathbf{c}_t)^{\frac{1-\beta}{2}}
+
\alpha \pi_{\mathrm{te}}(y_t \mid \mathbf{c}_t)^{\frac{1-\beta}{2}}
\right]^{\frac{2}{1-\beta}},
& \text{if }\beta \neq 1, \\[6pt]
\pi_\theta(y_t \mid \mathbf{c}_t)^{1-\alpha}
\pi_{\mathrm{te}}(y_t \mid \mathbf{c}_t)^\alpha,
& \text{if }\beta = 1.
\end{cases}
\end{equation}

The corresponding normalised assistant distribution is
\begin{equation}
\pi_{\mathrm{AMiD}}^{(\beta,\alpha)} (y_t \mid \mathbf{c}_t) = \frac{\widetilde{\pi}_{\mathrm{AMiD}}^{(\beta,\alpha)}(y_t \mid \mathbf{c}_t)}{Z_t},
\qquad
Z_t = \sum_{y'} \widetilde{\pi}_{\mathrm{AMiD}}^{(\beta,\alpha)} (y' \mid \mathbf{c}_t).
\end{equation}

This family recovers several commonly used interpolation schemes as special cases. In particular, setting $\beta=-1$ recovers the arithmetic (or convex) mixture, which we denote as \emph{ConMix}, whereas $\beta=1$ recovers the geometric mixture, which we denote as \emph{GeoMix}. Other values of $\beta$ define alternative interpolation geometries between the student and teacher distributions.

\subsection{Forward-KL Gradient Analysis} \label{app:amid_gradient}
We further analyse the forward-KL objective under the AMiD mixture. Following the gradient-coefficient analysis of \citet{kodistillm}, we consider the coefficient multiplying the student probability gradient. Therefore, at decoding step $t$, we apply a stop-gradient to the target distribution, yielding the following loss and gradient:
\begin{equation}
    \mathcal{L}_t
    =
    D_{\mathrm{KL}}
    \left(
    \operatorname{sg}\!\left[
    \pi_{\mathrm{AMiD}}^{(\beta,\alpha_t)}
    (\cdot\mid\mathbf{c}_t)
    \right]
    \,\middle\|\,
    \pi_\theta(\cdot\mid\mathbf{c}_t)
    \right),
\end{equation}
\begin{equation}
    \nabla_\theta \mathcal{L}_t
    =
    -\sum_{y}
    \frac{
        \pi_{\mathrm{AMiD}}^{(\beta,\alpha_t)}
        (y\mid\mathbf{c}_t)
    }{
        \pi_\theta(y\mid\mathbf{c}_t)
    }
    \nabla_\theta
    \pi_\theta(y\mid\mathbf{c}_t).
    \label{eq:amid_fkl_gradient}
\end{equation}
Thus, the ratio between the AMiD target and student probabilities determines the coefficient applied to each probability gradient.

\paragraph{General $\beta$-mixture.}
For $\beta\neq1$, substituting Equation~\ref{eq:amid} gives:
\begin{equation}
\frac{
    \pi_{\mathrm{AMiD}}^{(\beta,\alpha_t)}
    (y\mid\mathbf{c}_t)
}{
    \pi_\theta(y\mid\mathbf{c}_t)
}
=
\frac{1}{Z_t}
\left[
(1-\alpha_t)
+
\alpha_t
\left(
\frac{
    \pi_{\mathrm{te}}(y\mid\mathbf{c}_t)
}{
    \pi_\theta(y\mid\mathbf{c}_t)
}
\right)^{\frac{1-\beta}{2}}
\right]^{\frac{2}{1-\beta}}.
\label{eq:amid_gradient_coefficient}
\end{equation}

For $\beta<1$, the term inside the gradient coefficient interpolates between $1$ (when $\alpha_t=0$) and the teacher-to-student density ratio $\pi_{\mathrm{te}}/\pi_\theta$ (when $\alpha_t\rightarrow1$). Consequently, a smaller $\alpha_t$ can attenuate the contributions of a large probability mismatch.

\paragraph{Convex mixture ($\beta=-1$).}
For ConMix, Equation~\ref{eq:amid_gradient_coefficient} simplifies to:
\begin{equation}
\frac{
    \pi_{\mathrm{AMiD}}^{(-1,\alpha_t)}
    (y\mid\mathbf{c}_t)
}{
    \pi_\theta(y\mid\mathbf{c}_t)
}
=
(1-\alpha_t)
+
\alpha_t
\frac{
    \pi_{\mathrm{te}}(y\mid\mathbf{c}_t)
}{
    \pi_\theta(y\mid\mathbf{c}_t)
}.
\end{equation}

Here, $Z_t=1$, since any convex combination of two normalised distributions is itself normalised. Moreover, because $\sum_y \pi_\theta(y\mid\mathbf{c}t)=1$, we have $\sum_y \nabla\theta \pi_\theta(y\mid\mathbf{c}_t)=\nabla_\theta \sum_y \pi_\theta(y\mid\mathbf{c}_t)$. Thus, the contribution from the student component vanishes in Equation~\ref{eq:amid_fkl_gradient}, yielding

\begin{equation}
\nabla_\theta \mathcal{L}_t^{\mathrm{ConMix}}
=
\alpha_t
\nabla_\theta
D_{\mathrm{KL}}
\left(
\pi_{\mathrm{te}}(\cdot\mid\mathbf{c}_t)
\,\middle\|\,
\pi_\theta(\cdot\mid\mathbf{c}_t)
\right).
\label{eq:conmix_fkl_gradient}
\end{equation}

\paragraph{Geometric mixture ($\beta=1$).}
For GeoMix ($\beta=1$), the corresponding ratio is:
\begin{equation}
\frac{
    \pi_{\mathrm{AMiD}}^{(1,\alpha_t)}
    (y\mid\mathbf{c}_t)
}{
    \pi_\theta(y\mid\mathbf{c}_t)
}
=
\frac{1}{Z_t}
\left(
\frac{
    \pi_{\mathrm{te}}(y\mid\mathbf{c}_t)
}{
    \pi_\theta(y\mid\mathbf{c}_t)
}
\right)^{\alpha_t}.
\label{eq:geomix_gradient_coefficient}
\end{equation}
GeoMix raises the teacher-to-student probability ratio to the power \(\alpha_t\). When $\alpha_t<1$, extreme density ratios are compressed relative to direct teacher FKL, with stronger compression as $\alpha_t$ decreases.

\paragraph{Effect of Entropy-Aware Mixing.}
Suppose there is a confident conflict in which $\pi_\theta(y\mid\mathbf{c}_t) \ll \pi_{\mathrm{te}}(y\mid\mathbf{c}_t)$, so the teacher-to-student probability ratio is large. Under the standard FKL, this large ratio can dominate the update. Entropy-aware mixing instead downweights teacher influence when the student distribution has low entropy by using a small $\alpha_t$, keeping the mixed target closer to the student. As student entropy increases, $\alpha_t$ increases, allowing stronger teacher guidance. Thus, entropy-aware mixing adaptively moderates large student-teacher discrepancies rather than uniformly weakening supervision. Importantly, this interpretation uses entropy only to control the strength of the FKL update; it does not assume that entropy measures predictive correctness.

\newpage
\section{Implementation Details} \label{app:hyper_params}
\paragraph{Offline setup: Synthetic Trace Generation for SFT.}
Our offline setup consists of three stages: trace generation, trace verification, and supervised fine-tuning. Trace generation is implemented by extending vLLM speculative decoding, primarily through modifications to the rejection-sampling logic in \texttt{vllm/v1/sample/rejection\_sampler.py} and the associated speculative-decoding interfaces. This allows the sampler to access both student and teacher log-probabilities and, when enabled, replace the standard target with the proposed entropy-aware mixture during token acceptance and rejection sampling. Generated traces are then verified with rule-based rewards from \texttt{math-verify}, from which we compute the Pass@8 metric. The resulting verified traces are used to fine-tune the student with the \texttt{verl}~\citep{sheng2024hybridflow} SFT trainer. The corresponding hyperparameters are listed in Table~\ref{tab:hyperparams}.

\paragraph{Online setup: On-Policy Distillation.}
Implemented on top of \texttt{verl}~\citep{sheng2024hybridflow} by extending the distillation loss pipeline. In particular, we modify \texttt{verl/trainer/distillation} to support entropy-aware forward-KL and sampled reverse-KL objectives, expose convex and geometric mixtures as configurable targets, and compute the per-token mixing coefficient from student entropy during training. The same implementation also supports the online baselines FKL, OPD~\citep{lu2025onpolicydistillation}, EOPD~\citep{jin2026entropy}, Skew-KL~\citep{kodistillm}, and TAID~\citep{shing2025taid} within a unified framework. The corresponding hyperparameters are listed in Tables~\ref{tab:hyperparams}-\ref{tab:taid_hyperparams}. Please refer to the respective papers for details on the role of each hyperparameter.

\begin{table}[htbp]
\centering
\caption{Training hyperparameters for online and offline distillation.}
\label{tab:hyperparams}
\begin{tabular}{lcc}
\toprule
\textbf{Hyperparameter} & \textbf{Online} & \textbf{Offline} \\
\midrule
Optimizer & AdamW & AdamW \\
Learning rate & $3\times10^{-6}$ & $5\times10^{-6}$ \\
LR scheduler & cosine & cosine \\
Warmup ratio & 0.03 & 0.03 \\
Batch size & 128 & 128 \\
Samples per prompt & 1 & 1 (from 8 rollouts) \\
Temperature & 1.0 & 1.0 \\
Top-$p$ & 1.0 & 1.0 \\
Max response length & 4096 & 4096 \\
FKL top-$K$ & 64 & -- \\
Training epochs & 3 & 2 \\
\bottomrule
\end{tabular}
\end{table}

\begin{table}[htbp]
\centering
\caption{Entropy-Aware Mixing training hyperparameters.}
\label{tab:entropy_hyperparams}
\begin{tabular}{lcc}
\toprule
\textbf{Hyperparameter} & \textbf{Online} & \textbf{Offline} \\
\midrule
Teacher top-$K$ support & 64 & -- \\
Mixing transform \(f(x)\) & \(x^2\) & \(x^{1/4}\) \\
Student entropy top-$K$ & 32 & 32 \\
\bottomrule
\end{tabular}
\end{table}

\begin{table}[htbp]
\centering
\caption{EOPD training hyperparameters.}
\label{tab:eopd_hyperparams}
\begin{tabular}{lc}
\toprule
\textbf{Hyperparameter} & \textbf{Online} \\
\midrule
Teacher entropy top-$K$ & 32 \\
Teacher entropy threshold & 0.8 \\
\bottomrule
\end{tabular}
\end{table}

\begin{table}[htbp]
\centering
\caption{Skew-KL training hyperparameters.}
\label{tab:skew_kl_hyperparams}
\begin{tabular}{lc}
\toprule
\textbf{Hyperparameter} & \textbf{Online} \\
\midrule
$\alpha_t$ & 0.1 \\
\bottomrule
\end{tabular}
\end{table}

\begin{table}[htbp]
\centering
\caption{TAID training hyperparameters.}
\label{tab:taid_hyperparams}
\begin{tabular}{lc}
\toprule
\textbf{Hyperparameter} & \textbf{Online} \\
\midrule
$t_{\mathrm{start}}$ & 0.3 \\
$t_{\mathrm{end}}$ & 1.0 \\
Learning rate of $t$ & $5 \times 10^{-4}$ \\
$\beta$ (momentum coefficient) & 0.99 \\
Adaptive TAID & enabled \\
\bottomrule
\end{tabular}
\end{table}

\paragraph{Chat Template.}
For Qwen3 models in non-thinking mode, we use the following chat template to encourage step-by-step reasoning and format the final answer within \texttt{\textbackslash boxed\{\}}.

\begin{center}
\begin{tcolorbox}[colback=gray!5!white, colframe=gray!50!black, width=0.6\textwidth, sharp corners]
\begin{verbatim}
<|im_start|>user
{prompt}
Let's think step by step and output the 
final answer within \boxed{}.<|im_end|>
<|im_start|>assistant
<think>

</think>
\end{verbatim}
\end{tcolorbox}
\end{center}

\begin{figure}[htbp]
    \centering
    \begin{subfigure}[b]{0.32\textwidth}
        \centering
        \includegraphics[width=\textwidth]{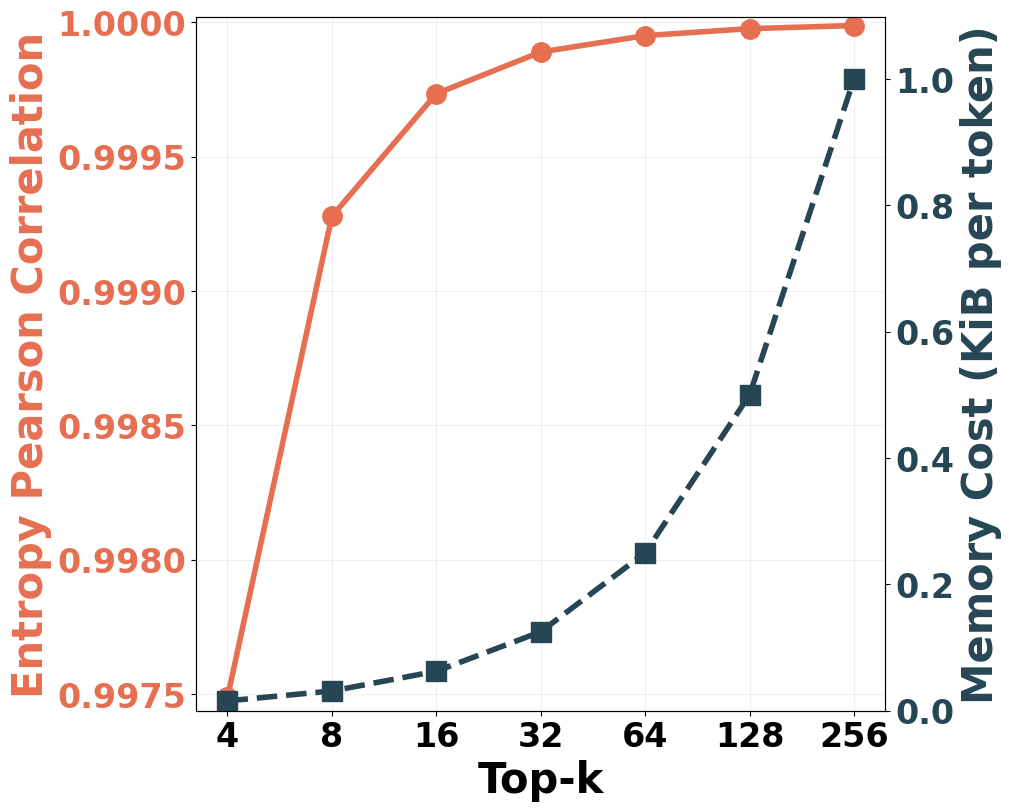}
        \caption{Qwen3-1.7B-Base entropy on Qwen3-8B generations (MATH).}
    \end{subfigure}
    \hfill
    \begin{subfigure}[b]{0.32\textwidth}
        \centering
        \includegraphics[width=\textwidth]{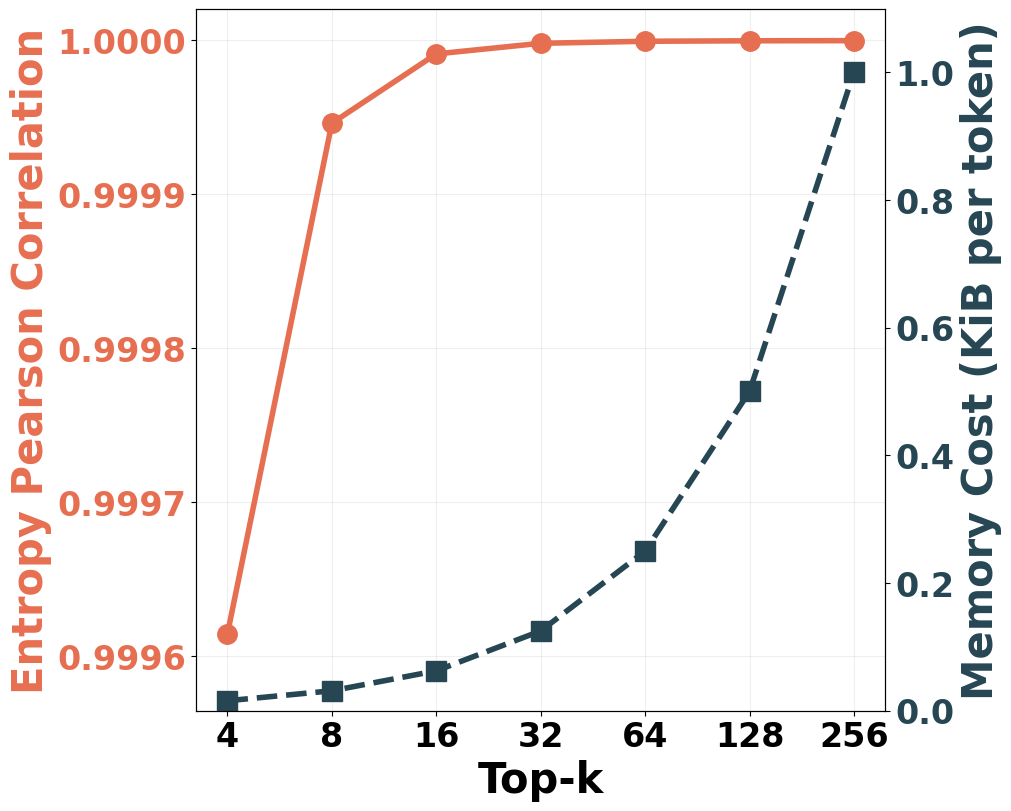}
        \caption{Qwen3-1.7B entropy on Qwen3-8B generations (MATH).}
    \end{subfigure}
    \hfill
    \begin{subfigure}[b]{0.3\textwidth}
        \centering
        \includegraphics[width=\textwidth]{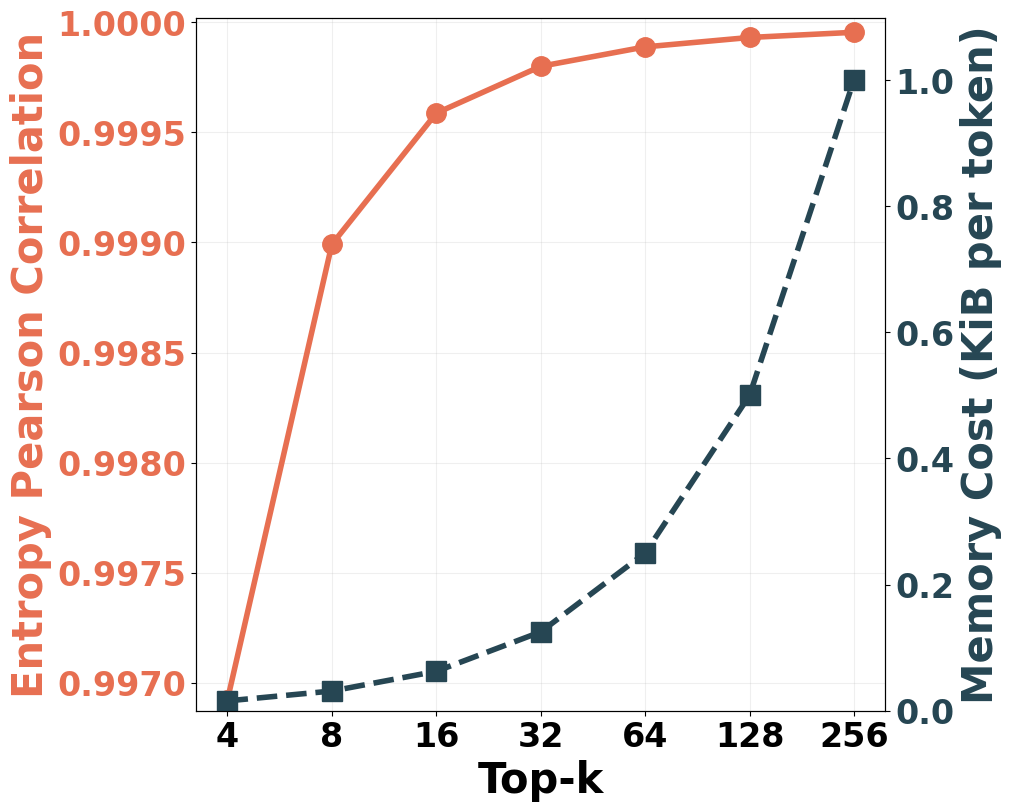}
        \caption{Granite-3.2-2B-Instruct entropy on Granite-3.3-8B-Instruct generations (DAPO-Math-14k).}
    \end{subfigure}
    \caption{Correlation between the top-$K$ entropy approximation and the exact full-vocabulary entropy across different student-teacher pairs.}
    \label{fig:entropy_approx}
\end{figure}

\paragraph{Adaptive Gating Signal.}
The adaptive gating signal introduced in Section~\ref{sec:entropy_aware_mixing} uses the student’s predictive entropy to interpolate between the student and teacher distributions. Computing the exact entropy over the full vocabulary is unnecessary for this purpose and incurs avoidable computational cost. We therefore approximate the entropy using only the top-$K$ logits, substantially reducing both compute and memory overhead. Figure~\ref{fig:entropy_approx} shows that this top-$K$ approximation remains very strongly correlated with the exact full-vocabulary entropy across all student-teacher pairs considered. While \citet{diao2026entropy} use $k=20$, we use $k=32$, which provides a better trade-off, yielding consistently stronger correlation across model families while retaining a small memory footprint.

\begin{figure}
    \centering
    \includegraphics[width=0.6\linewidth]{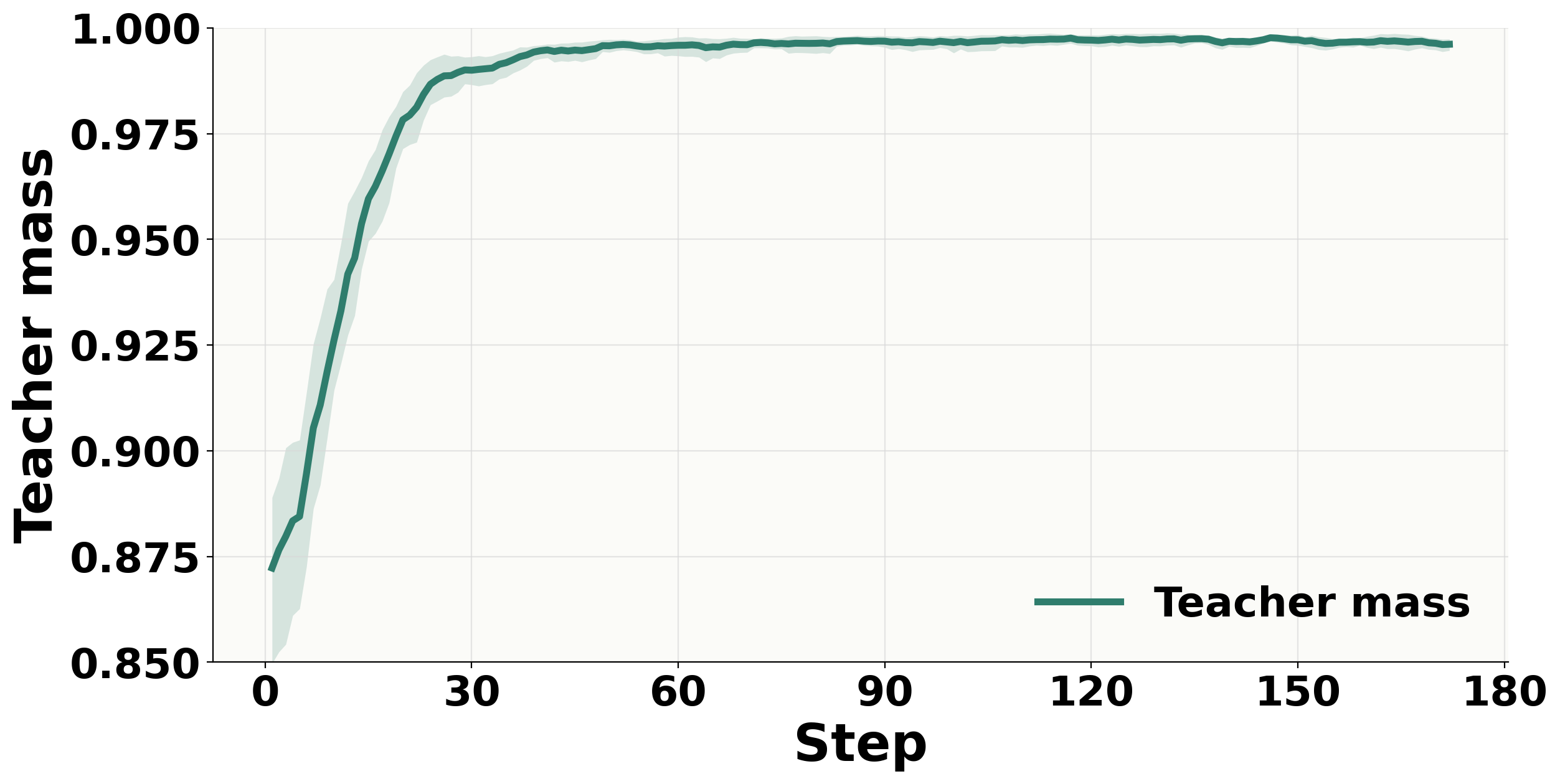}
    \caption{Teacher probability mass captured by the top-64 tokens during on-policy distillation. The combined probability mass of the top-64 covers most of the teacher distribution throughout training and quickly approaches full mass.}
    \label{fig:teacher_mass}
\end{figure}

\paragraph{Teacher support.}
We use the teacher's top-$K$ support during on-policy distillation, since extracting full-vocabulary log-probabilities is substantially more expensive. Figure~\ref{fig:teacher_mass} shows that the teacher's top-64 tokens already capture $\geq 90\%$ of the probability mass early in training and rapidly approach nearly full mass. We therefore perform mixing over the renormalised top-64 teacher distribution, which provides an efficient approximation to full-distribution mixing. We leave a systematic study of the effect of $K$ and the resulting top-$K$ approximation error to future work.

\newpage
\section{Results: Confident Student Errors} \label{app:confident_conflicts}
A potential limitation of Entropy-Aware Mixing is that low student entropy does not necessarily imply correctness. In particular, the student may be confident but wrong, and our entropy-based interpolation coefficient would reduce teacher influence. We therefore evaluate whether confident student errors lead to a systematic failure of entropy-aware mixing.

Using the offline distillation experimental setup described in Section~\ref{sec:experiment}, we first group Qwen3-1.7B generated trajectories by their average entropy. We observe that responses in the lowest-entropy quartile are substantially more accurate than the rest ($92.82\%$ vs. $64.06\%$), suggesting that higher-entropy sequences may require additional inference-time compute to benefit from better exploration. 

\paragraph{Prompt-level analysis.}
We filter trajectories in the lowest-entropy quartile that contain at least one incorrect rollout, yielding 542 prompts. For each prompt, we report the Avg@8 and Pass@8 metrics for the student-generated, teacher-generated, ConMix-SD, and GeoMix-SD distributions.

\begin{table}[H]
\centering
\caption{Performance on MATH prompts with at least one incorrect low-entropy student rollout.}
\label{tab:confident-error-prompt}
\begin{tabular}{lcc}
\toprule
\textbf{Method} & \textbf{Avg@8} & \textbf{Pass@8} \\
\midrule
Student-generated (Qwen3-1.7B) & 47.6 & 81.7 \\
Teacher-generated (Qwen3-8B) & 77.7 & 91.9 \\
ConMix-SD & 77.3 & 91.9 \\
GeoMix-SD & 77.2 & 92.6 \\
\bottomrule
\end{tabular}
\end{table}

Both entropy-aware methods substantially outperform student-only sampling and achieve performance comparable to teacher sampling. This suggests that the presence of confident student errors does not, by itself, prevent entropy-aware mixtures from generating correct trajectories.

\paragraph{Prefix-level analysis.}
To test whether the mixtures can recover from confident conflicts, we select incorrect trajectories from the lowest-entropy quartile, where the average entropy is low. For each trajectory, we retain the prefix preceding the first confident conflict and resample continuations from the student, teacher, ConMix-SD, and GeoMix-SD distributions.

\begin{table}[H]
\centering
\caption{Prefix-level intervention from incorrect low-entropy student trajectories.}
\label{tab:confident-error-prefix}
\begin{tabular}{lc}
\toprule
\textbf{Method} & \textbf{Mean success} \\
\midrule
Student-generated (Qwen3-1.7B) & 34.7 \\
Teacher-generated (Qwen3-8B) & 57.2 \\
ConMix-SD & 56.7 \\
GeoMix-SD & 56.3 \\
\bottomrule
\end{tabular}
\end{table}

ConMix-SD and GeoMix-SD again substantially outperform student sampling and achieve success rates close to teacher sampling. The non-zero success rate of the student is expected because all methods resample from the same prefix, which may still result in a correct continuation. These results show that entropy-aware mixing can recover from prefixes associated with confident student errors, despite assigning low teacher influence in low-entropy states.

\newpage
\section{Results: Additional Qwen3 Experiments} \label{app:qwen3_app_results}
\subsection{Mathematical Reasoning} \label{app:qwen3_results}
This appendix reports the complete mathematical reasoning results for Qwen3. While Section~\ref{sec:experiment} presents a representative subset for readability, Tables~\ref{tab:qwen3_17b_math_sft}–\ref{tab:qwen3_06b_base_math_kd} provide the full results for both offline and online distillation across mathematical benchmarks (AIME24, AIME25, AMC23, GSM8K, MATH500, and Minerva) and general-capability benchmarks (HellaSwag, IFEval, and MMLU). These results corroborate the main findings and demonstrate that the observed trends hold consistently across the full benchmark suite.

\begin{table*}[h]
\centering
\caption{Offline distillation results on Qwen3-1.7B using MATH. Results are reported as mean$_{\pm \text{std}}$ across three seeds. \textbf{Bold} marks the best result and \underline{underlining} marks the second-best result; ties receive the same formatting.}
\label{tab:qwen3_17b_math_sft}
\resizebox{\textwidth}{!}{%
\begin{tabular}{ll cccccc}
\toprule
\multirow{2}{*}{\textbf{Benchmark}}
& \multirow{2}{*}{\textbf{Metric}}
& \multicolumn{6}{c}{\textbf{Qwen3-1.7B} (MATH)} \\
\cmidrule(lr){3-8}
&
& \textbf{\makecell{Student-\\generated}}
& \textbf{\makecell{Teacher-\\generated}}
& \textbf{SKD}
& \textbf{EAFT}
& \cellcolor{blue!10}\textbf{ConMix-SFT}
& \cellcolor{blue!10}\textbf{GeoMix-SFT} \\
\midrule
\multicolumn{8}{l}{\textbf{\textit{Mathematical reasoning}}} \\

\midrule
\multirow{3}{*}{AIME24}
& Avg@64
& \underline{11.06}$_{\pm 0.63}$
& 8.70$_{\pm 0.42}$
& 9.49$_{\pm 0.53}$
& 10.17$_{\pm 0.30}$
& \cellcolor{blue!10}10.57$_{\pm 0.66}$
& \cellcolor{blue!10}\textbf{11.54}$_{\pm 0.55}$ \\
& Pass@8
& 24.97$_{\pm 0.06}$
& 23.96$_{\pm 1.75}$
& \underline{26.47}$_{\pm 1.32}$
& 26.07$_{\pm 2.17}$
& \cellcolor{blue!10}25.35$_{\pm 2.98}$
& \cellcolor{blue!10}\textbf{27.59}$_{\pm 0.27}$ \\
& Pass@64
& 41.67$_{\pm 2.35}$
& 46.67$_{\pm 5.09}$
& \underline{47.78}$_{\pm 5.09}$
& 45.56$_{\pm 5.09}$
& \cellcolor{blue!10}43.34$_{\pm 9.43}$
& \cellcolor{blue!10}\textbf{50.00}$_{\pm 0.00}$ \\

\midrule
\multirow{3}{*}{AIME25}
& Avg@64
& 9.29$_{\pm 1.00}$
& 8.44$_{\pm 0.47}$
& 8.80$_{\pm 0.46}$
& 8.73$_{\pm 0.47}$
& \cellcolor{blue!10}\underline{9.53}$_{\pm 1.25}$
& \cellcolor{blue!10}\textbf{9.71}$_{\pm 0.85}$ \\
& Pass@8
& 24.13$_{\pm 1.03}$
& 22.27$_{\pm 0.85}$
& \underline{25.02}$_{\pm 1.34}$
& 23.81$_{\pm 0.36}$
& \cellcolor{blue!10}\textbf{25.86}$_{\pm 2.28}$
& \cellcolor{blue!10}24.58$_{\pm 0.59}$ \\
& Pass@64
& 40.00$_{\pm 0.00}$
& 36.67$_{\pm 4.56}$
& \underline{43.33}$_{\pm 3.34}$
& 36.67$_{\pm 5.77}$
& \cellcolor{blue!10}\textbf{51.67}$_{\pm 2.35}$
& \cellcolor{blue!10}42.22$_{\pm 4.71}$ \\

\midrule
\multirow{3}{*}{AMC23}
& Avg@64
& 41.57$_{\pm 0.72}$
& 39.88$_{\pm 0.84}$
& 42.25$_{\pm 0.75}$
& 40.78$_{\pm 0.93}$
& \cellcolor{blue!10}\textbf{43.07}$_{\pm 1.19}$
& \cellcolor{blue!10}\underline{42.36}$_{\pm 0.91}$ \\
& Pass@8
& 76.09$_{\pm 1.07}$
& 76.55$_{\pm 1.12}$
& \textbf{77.60}$_{\pm 1.31}$
& \underline{77.54}$_{\pm 0.92}$
& \cellcolor{blue!10}77.10$_{\pm 1.48}$
& \cellcolor{blue!10}75.72$_{\pm 1.81}$ \\
& Pass@64
& 91.67$_{\pm 3.82}$
& 90.00$_{\pm 1.97}$
& \textbf{94.17}$_{\pm 1.44}$
& 92.50$_{\pm 2.50}$
& \cellcolor{blue!10}92.50$_{\pm 3.54}$
& \cellcolor{blue!10}\underline{93.75}$_{\pm 1.77}$ \\

\midrule
\multirow{3}{*}{GSM8K}
& Avg@64
& 80.94$_{\pm 0.10}$
& 79.98$_{\pm 0.20}$
& 79.73$_{\pm 0.23}$
& 79.67$_{\pm 0.16}$
& \cellcolor{blue!10}\textbf{81.12}$_{\pm 0.37}$
& \cellcolor{blue!10}\underline{81.03}$_{\pm 0.16}$ \\
& Pass@8
& 93.12$_{\pm 0.14}$
& \underline{94.45}$_{\pm 0.14}$
& 94.23$_{\pm 0.25}$
& 94.23$_{\pm 0.03}$
& \cellcolor{blue!10}\textbf{94.65}$_{\pm 0.37}$
& \cellcolor{blue!10}94.37$_{\pm 0.00}$ \\
& Pass@64
& 97.07$_{\pm 0.23}$
& \textbf{98.20}$_{\pm 0.26}$
& \underline{98.00}$_{\pm 0.20}$
& 97.93$_{\pm 0.31}$
& \cellcolor{blue!10}\underline{98.00}$_{\pm 0.28}$
& \cellcolor{blue!10}\textbf{98.20}$_{\pm 0.28}$ \\

\midrule
\multirow{3}{*}{MATH500}
& Avg@64
& 70.56$_{\pm 0.09}$
& 69.36$_{\pm 0.13}$
& 69.36$_{\pm 0.12}$
& 69.16$_{\pm 0.13}$
& \cellcolor{blue!10}\textbf{70.99}$_{\pm 0.44}$
& \cellcolor{blue!10}\underline{70.90}$_{\pm 0.49}$ \\
& Pass@8
& 88.76$_{\pm 0.38}$
& 89.45$_{\pm 0.10}$
& \underline{89.55}$_{\pm 0.08}$
& 89.31$_{\pm 0.12}$
& \cellcolor{blue!10}\textbf{90.00}$_{\pm 0.24}$
& \cellcolor{blue!10}89.47$_{\pm 0.18}$ \\
& Pass@64
& 95.33$_{\pm 0.64}$
& 96.00$_{\pm 0.61}$
& 95.47$_{\pm 0.61}$
& \underline{96.13}$_{\pm 0.61}$
& \cellcolor{blue!10}\textbf{96.30}$_{\pm 0.14}$
& \cellcolor{blue!10}95.40$_{\pm 0.28}$ \\

\midrule
\multirow{3}{*}{Minerva}
& Avg@64
& \underline{27.94}$_{\pm 0.10}$
& 27.34$_{\pm 0.26}$
& 27.58$_{\pm 0.37}$
& 27.14$_{\pm 0.14}$
& \cellcolor{blue!10}27.84$_{\pm 0.25}$
& \cellcolor{blue!10}\textbf{27.97}$_{\pm 0.11}$ \\
& Pass@8
& 43.85$_{\pm 0.54}$
& 44.74$_{\pm 0.25}$
& 45.10$_{\pm 0.21}$
& 45.35$_{\pm 0.29}$
& \cellcolor{blue!10}\underline{46.38}$_{\pm 0.02}$
& \cellcolor{blue!10}\textbf{46.52}$_{\pm 0.15}$ \\
& Pass@64
& 54.04$_{\pm 0.97}$
& 54.78$_{\pm 1.14}$
& 56.62$_{\pm 0.98}$
& \underline{56.86}$_{\pm 1.29}$
& \cellcolor{blue!10}\textbf{56.99}$_{\pm 0.00}$
& \cellcolor{blue!10}56.80$_{\pm 0.78}$ \\

\midrule
\multirow{3}{*}{\textbf{Average}}
& Avg@64
& 40.23$_{\pm 0.23}$
& 38.95$_{\pm 0.39}$
& 39.54$_{\pm 0.19}$
& 39.28$_{\pm 0.19}$
& \cellcolor{blue!10}\underline{40.52}$_{\pm 0.33}$
& \cellcolor{blue!10}\textbf{40.59}$_{\pm 0.24}$ \\
& Pass@8
& 58.49$_{\pm 0.27}$
& 58.57$_{\pm 0.70}$
& 59.66$_{\pm 0.39}$
& 59.39$_{\pm 0.40}$
& \cellcolor{blue!10}\textbf{59.89}$_{\pm 0.68}$
& \cellcolor{blue!10}\underline{59.71}$_{\pm 0.32}$ \\
& Pass@64
& 69.96$_{\pm 0.77}$
& 70.39$_{\pm 2.27}$
& 72.56$_{\pm 1.06}$
& 70.94$_{\pm 1.37}$
& \cellcolor{blue!10}\textbf{73.13}$_{\pm 1.72}$
& \cellcolor{blue!10}\underline{72.73}$_{\pm 0.85}$ \\

\midrule
\multicolumn{8}{l}{\textbf{\textit{General capabilities}}} \\
\midrule

HellaSwag
& Avg@8
& 50.59$_{\pm 0.03}$
& 51.59$_{\pm 0.32}$
& 51.76$_{\pm 0.13}$
& 51.91$_{\pm 0.32}$
& \cellcolor{blue!10}\textbf{55.51}$_{\pm 0.68}$
& \cellcolor{blue!10}\underline{55.07}$_{\pm 0.66}$ \\

IFEval
& Avg@8
& \underline{41.71}$_{\pm 0.65}$
& 38.82$_{\pm 1.19}$
& 38.32$_{\pm 1.80}$
& 33.76$_{\pm 0.83}$
& \cellcolor{blue!10}41.50$_{\pm 3.88}$
& \cellcolor{blue!10}\textbf{41.96}$_{\pm 1.57}$ \\

MMLU
& Avg@8
& \textbf{63.30}$_{\pm 0.20}$
& 61.00$_{\pm 0.97}$
& 61.40$_{\pm 0.70}$
& 60.80$_{\pm 0.50}$
& \cellcolor{blue!10}62.20$_{\pm 0.50}$
& \cellcolor{blue!10}\underline{62.70}$_{\pm 0.40}$ \\

\bottomrule
\end{tabular}%
}
\end{table*}

\begin{table*}[h]
\centering
\caption{Online distillation results on Qwen3-1.7B-Base using MATH. Results are reported as mean$_{\pm \text{std}}$ across three seeds. \textbf{Bold} marks the best result and \underline{underlining} marks the second-best result; ties receive the same formatting.}
\label{tab:qwen3_17b_base_math_kd}
\resizebox{\textwidth}{!}{%
\begin{tabular}{ll ccccccc}
\toprule
\multirow{2}{*}{\textbf{Benchmark}}
& \multirow{2}{*}{\textbf{Metric}}
& \multicolumn{7}{c}{\textbf{Qwen3-1.7B-Base} (MATH)} \\
\cmidrule(lr){3-9}
&
& \textbf{FKL}
& \textbf{OPD}
& \textbf{EOPD}
& \textbf{Skew-KL}
& \textbf{TAID}
& \cellcolor{blue!10}\textbf{ConMix-FKL}
& \cellcolor{blue!10}\textbf{GeoMix-FKL} \\
\midrule
\multicolumn{9}{l}{\textbf{\textit{Mathematical reasoning}}} \\
\midrule

\multirow{3}{*}{AIME24}
& Avg@64
& 5.28$_{\pm 0.44}$
& \underline{7.63}$_{\pm 0.26}$
& 7.20$_{\pm 0.43}$
& 4.91$_{\pm 0.11}$
& 5.89$_{\pm 0.42}$
& \cellcolor{blue!10}6.75$_{\pm 0.59}$
& \cellcolor{blue!10}\textbf{8.15}$_{\pm 0.27}$ \\
& Pass@8
& 18.17$_{\pm 0.73}$
& 21.15$_{\pm 0.85}$
& 19.96$_{\pm 1.02}$
& 17.25$_{\pm 0.62}$
& 18.41$_{\pm 2.68}$
& \cellcolor{blue!10}\underline{21.37}$_{\pm 0.38}$
& \cellcolor{blue!10}\textbf{22.67}$_{\pm 0.95}$ \\
& Pass@64
& 35.56$_{\pm 1.93}$
& 35.57$_{\pm 1.77}$
& 34.44$_{\pm 1.93}$
& \underline{36.67}$_{\pm 6.67}$
& \underline{36.67}$_{\pm 3.34}$
& \cellcolor{blue!10}35.55$_{\pm 3.85}$
& \cellcolor{blue!10}\textbf{39.72}$_{\pm 2.87}$ \\

\midrule
\multirow{3}{*}{AIME25}
& Avg@64
& 3.87$_{\pm 0.42}$
& 4.87$_{\pm 0.10}$
& 4.67$_{\pm 0.08}$
& 3.89$_{\pm 0.51}$
& 4.32$_{\pm 0.47}$
& \cellcolor{blue!10}\underline{4.88}$_{\pm 0.30}$
& \cellcolor{blue!10}\textbf{5.13}$_{\pm 0.15}$ \\
& Pass@8
& 14.56$_{\pm 1.91}$
& 16.27$_{\pm 0.44}$
& 15.96$_{\pm 0.84}$
& 15.37$_{\pm 1.78}$
& 15.79$_{\pm 1.74}$
& \cellcolor{blue!10}\textbf{18.43}$_{\pm 1.75}$
& \cellcolor{blue!10}\underline{17.65}$_{\pm 0.46}$ \\
& Pass@64
& 34.44$_{\pm 1.93}$
& 33.06$_{\pm 1.60}$
& 35.56$_{\pm 1.93}$
& 31.11$_{\pm 2.85}$
& 33.34$_{\pm 5.77}$
& \cellcolor{blue!10}\textbf{37.78}$_{\pm 5.09}$
& \cellcolor{blue!10}\underline{37.70}$_{\pm 1.68}$ \\

\midrule
\multirow{3}{*}{AMC23}
& Avg@64
& 31.39$_{\pm 0.77}$
& \underline{37.00}$_{\pm 0.43}$
& 35.32$_{\pm 0.49}$
& 30.05$_{\pm 0.80}$
& 32.67$_{\pm 0.79}$
& \cellcolor{blue!10}34.58$_{\pm 0.34}$
& \cellcolor{blue!10}\textbf{37.32}$_{\pm 0.08}$ \\
& Pass@8
& 66.17$_{\pm 1.77}$
& \underline{70.77}$_{\pm 0.65}$
& 69.15$_{\pm 0.45}$
& 66.23$_{\pm 1.37}$
& 67.79$_{\pm 0.59}$
& \cellcolor{blue!10}68.73$_{\pm 1.16}$
& \cellcolor{blue!10}\textbf{70.92}$_{\pm 0.17}$ \\
& Pass@64
& \textbf{92.50}$_{\pm 2.50}$
& 92.08$_{\pm 1.58}$
& 85.00$_{\pm 2.50}$
& 91.67$_{\pm 3.82}$
& 89.17$_{\pm 3.82}$
& \cellcolor{blue!10}89.17$_{\pm 5.77}$
& \cellcolor{blue!10}\underline{92.09}$_{\pm 0.83}$ \\

\midrule
\multirow{3}{*}{GSM8K}
& Avg@64
& 79.22$_{\pm 0.12}$
& \underline{81.57}$_{\pm 0.10}$
& 81.55$_{\pm 0.08}$
& 79.10$_{\pm 0.05}$
& 80.02$_{\pm 0.14}$
& \cellcolor{blue!10}80.36$_{\pm 0.04}$
& \cellcolor{blue!10}\textbf{81.62}$_{\pm 0.04}$ \\
& Pass@8
& 95.48$_{\pm 0.22}$
& 95.37$_{\pm 0.08}$
& \textbf{95.72}$_{\pm 0.17}$
& 95.42$_{\pm 0.04}$
& 95.58$_{\pm 0.15}$
& \cellcolor{blue!10}95.54$_{\pm 0.07}$
& \cellcolor{blue!10}\underline{95.59}$_{\pm 0.15}$ \\
& Pass@64
& \textbf{99.47}$_{\pm 0.12}$
& 99.12$_{\pm 0.15}$
& 99.13$_{\pm 0.12}$
& 99.33$_{\pm 0.23}$
& 99.07$_{\pm 0.42}$
& \cellcolor{blue!10}99.27$_{\pm 0.31}$
& \cellcolor{blue!10}\underline{99.41}$_{\pm 0.08}$ \\

\midrule
\multirow{3}{*}{MATH500}
& Avg@64
& 59.68$_{\pm 0.24}$
& \underline{64.43}$_{\pm 0.11}$
& 63.06$_{\pm 0.07}$
& 57.31$_{\pm 0.16}$
& 60.47$_{\pm 0.07}$
& \cellcolor{blue!10}62.13$_{\pm 0.07}$
& \cellcolor{blue!10}\textbf{64.45}$_{\pm 0.15}$ \\
& Pass@8
& 84.65$_{\pm 0.39}$
& \underline{85.98}$_{\pm 0.04}$
& 85.40$_{\pm 0.05}$
& 83.61$_{\pm 0.04}$
& 84.78$_{\pm 0.25}$
& \cellcolor{blue!10}85.66$_{\pm 0.07}$
& \cellcolor{blue!10}\textbf{86.04}$_{\pm 0.08}$ \\
& Pass@64
& \underline{94.13}$_{\pm 0.31}$
& 93.79$_{\pm 0.14}$
& 93.47$_{\pm 0.61}$
& 93.40$_{\pm 1.06}$
& \textbf{94.27}$_{\pm 0.23}$
& \cellcolor{blue!10}94.07$_{\pm 1.29}$
& \cellcolor{blue!10}93.95$_{\pm 0.22}$ \\

\midrule
\multirow{3}{*}{Minerva}
& Avg@64
& 23.68$_{\pm 0.20}$
& \textbf{24.81}$_{\pm 0.04}$
& 24.71$_{\pm 0.13}$
& 22.65$_{\pm 0.34}$
& 23.48$_{\pm 0.23}$
& \cellcolor{blue!10}24.29$_{\pm 0.38}$
& \cellcolor{blue!10}\underline{24.79}$_{\pm 0.11}$ \\
& Pass@8
& \underline{44.90}$_{\pm 0.20}$
& 44.23$_{\pm 0.11}$
& 44.84$_{\pm 0.44}$
& 44.09$_{\pm 0.26}$
& 43.96$_{\pm 0.17}$
& \cellcolor{blue!10}\underline{44.90}$_{\pm 0.69}$
& \cellcolor{blue!10}\textbf{45.17}$_{\pm 0.28}$ \\
& Pass@64
& \textbf{58.21}$_{\pm 1.18}$
& 55.96$_{\pm 0.28}$
& 57.47$_{\pm 1.29}$
& 57.35$_{\pm 1.11}$
& 57.11$_{\pm 0.76}$
& \cellcolor{blue!10}57.60$_{\pm 1.18}$
& \cellcolor{blue!10}\underline{58.05}$_{\pm 0.23}$ \\

\midrule
\multirow{3}{*}{\textbf{Average}}
& Avg@64
& 33.85$_{\pm 0.36}$
& \underline{36.72}$_{\pm 0.09}$
& 36.09$_{\pm 0.21}$
& 32.99$_{\pm 0.33}$
& 34.48$_{\pm 0.35}$
& \cellcolor{blue!10}35.50$_{\pm 0.29}$
& \cellcolor{blue!10}\textbf{36.91}$_{\pm 0.06}$ \\
& Pass@8
& 53.99$_{\pm 0.87}$
& 55.63$_{\pm 0.19}$
& 55.17$_{\pm 0.49}$
& 53.66$_{\pm 0.69}$
& 54.39$_{\pm 0.93}$
& \cellcolor{blue!10}\underline{55.77}$_{\pm 0.69}$
& \cellcolor{blue!10}\textbf{56.34}$_{\pm 0.19}$ \\
& Pass@64
& \underline{69.05}$_{\pm 1.33}$
& 68.26$_{\pm 0.48}$
& 67.51$_{\pm 1.40}$
& 68.26$_{\pm 2.62}$
& 68.27$_{\pm 2.39}$
& \cellcolor{blue!10}68.91$_{\pm 2.92}$
& \cellcolor{blue!10}\textbf{70.17}$_{\pm 0.57}$ \\

\midrule
\multicolumn{9}{l}{\textbf{\textit{General capabilities}}} \\
\midrule

HellaSwag
& Avg@8
& 61.33$_{\pm 0.07}$
& 61.18$_{\pm 0.03}$
& \underline{61.39}$_{\pm 0.04}$
& 61.10$_{\pm 0.07}$
& 61.28$_{\pm 0.06}$
& \cellcolor{blue!10}61.34$_{\pm 0.03}$
& \cellcolor{blue!10}\textbf{61.42}$_{\pm 0.02}$ \\

IFEval
& Avg@8
& 23.48$_{\pm 0.49}$
& 24.21$_{\pm 1.12}$
& 24.46$_{\pm 1.57}$
& 23.41$_{\pm 1.20}$
& 24.28$_{\pm 1.76}$
& \cellcolor{blue!10}\underline{24.95}$_{\pm 0.74}$
& \cellcolor{blue!10}\textbf{25.26}$_{\pm 0.77}$ \\

MMLU
& Avg@8
& 47.90$_{\pm 0.40}$
& 47.80$_{\pm 1.10}$
& 48.20$_{\pm 0.40}$
& 46.50$_{\pm 0.30}$
& 48.00$_{\pm 0.20}$
& \cellcolor{blue!10}\underline{48.90}$_{\pm 0.40}$
& \cellcolor{blue!10}\textbf{49.30}$_{\pm 0.20}$ \\

\bottomrule
\end{tabular}%
}
\end{table*}

\begin{table*}[h]
\centering
\caption{Offline distillation results on Qwen3-0.6B using MATH. Results are reported as mean$_{\pm \text{std}}$ across three seeds. \textbf{Bold} marks the best result and \underline{underlining} marks the second-best result; ties receive the same formatting.}
\label{tab:qwen3_06b_math_sft}
\resizebox{\textwidth}{!}{%
\begin{tabular}{ll cccccc}
\toprule
\multirow{2}{*}{\textbf{Benchmark}}
& \multirow{2}{*}{\textbf{Metric}}
& \multicolumn{6}{c}{\textbf{Qwen3-0.6B} (MATH)} \\
\cmidrule(lr){3-8}
&
& \textbf{\makecell{Student-\\generated}}
& \textbf{\makecell{Teacher-\\generated}}
& \textbf{SKD}
& \textbf{EAFT}
& \cellcolor{blue!10}\textbf{ConMix-SFT}
& \cellcolor{blue!10}\textbf{GeoMix-SFT} \\
\midrule
\multicolumn{8}{l}{\textbf{\textit{Mathematical reasoning}}} \\
\midrule

\multirow{3}{*}{AIME24}
& Avg@64
& 1.79$_{\pm 0.26}$
& 2.66$_{\pm 0.45}$
& \underline{2.74}$_{\pm 0.17}$
& 2.40$_{\pm 0.23}$
& \cellcolor{blue!10}2.71$_{\pm 0.18}$
& \cellcolor{blue!10}\textbf{2.90}$_{\pm 0.43}$ \\
& Pass@8
& 6.86$_{\pm 1.01}$
& \textbf{10.47}$_{\pm 0.87}$
& \underline{10.22}$_{\pm 1.46}$
& 9.09$_{\pm 1.32}$
& \cellcolor{blue!10}9.51$_{\pm 0.41}$
& \cellcolor{blue!10}9.37$_{\pm 1.14}$ \\
& Pass@64
& 23.33$_{\pm 3.34}$
& \underline{27.78}$_{\pm 1.92}$
& \textbf{31.11}$_{\pm 3.85}$
& 26.67$_{\pm 8.82}$
& \cellcolor{blue!10}22.22$_{\pm 1.92}$
& \cellcolor{blue!10}26.67$_{\pm 6.67}$ \\

\midrule
\multirow{3}{*}{AIME25}
& Avg@64
& 2.33$_{\pm 0.46}$
& 3.63$_{\pm 0.30}$
& 3.85$_{\pm 0.41}$
& 3.63$_{\pm 0.21}$
& \cellcolor{blue!10}\underline{4.11}$_{\pm 0.32}$
& \cellcolor{blue!10}\textbf{4.17}$_{\pm 0.41}$ \\
& Pass@8
& 10.21$_{\pm 0.95}$
& 14.08$_{\pm 0.58}$
& 14.76$_{\pm 1.11}$
& \underline{14.94}$_{\pm 0.97}$
& \cellcolor{blue!10}\textbf{15.07}$_{\pm 0.73}$
& \cellcolor{blue!10}14.75$_{\pm 1.96}$ \\
& Pass@64
& 22.22$_{\pm 3.85}$
& \underline{27.78}$_{\pm 5.09}$
& \textbf{28.89}$_{\pm 5.09}$
& \underline{27.78}$_{\pm 5.09}$
& \cellcolor{blue!10}\underline{27.78}$_{\pm 1.92}$
& \cellcolor{blue!10}26.67$_{\pm 8.82}$ \\

\midrule
\multirow{3}{*}{AMC23}
& Avg@64
& 23.31$_{\pm 0.61}$
& \textbf{28.27}$_{\pm 0.69}$
& \underline{28.22}$_{\pm 0.45}$
& 27.06$_{\pm 0.62}$
& \cellcolor{blue!10}28.03$_{\pm 0.82}$
& \cellcolor{blue!10}27.89$_{\pm 0.67}$ \\
& Pass@8
& 54.32$_{\pm 0.74}$
& 61.31$_{\pm 2.67}$
& 61.69$_{\pm 0.36}$
& \textbf{62.05}$_{\pm 1.04}$
& \cellcolor{blue!10}61.39$_{\pm 0.54}$
& \cellcolor{blue!10}\underline{61.91}$_{\pm 1.22}$ \\
& Pass@64
& 77.50$_{\pm 2.50}$
& \textbf{86.67}$_{\pm 3.82}$
& \underline{84.17}$_{\pm 3.82}$
& \underline{84.17}$_{\pm 3.82}$
& \cellcolor{blue!10}80.83$_{\pm 1.44}$
& \cellcolor{blue!10}\textbf{86.67}$_{\pm 6.29}$ \\

\midrule
\multirow{3}{*}{GSM8K}
& Avg@64
& 59.41$_{\pm 0.25}$
& 60.09$_{\pm 0.15}$
& 60.13$_{\pm 0.25}$
& 59.07$_{\pm 0.16}$
& \cellcolor{blue!10}\underline{62.94}$_{\pm 0.51}$
& \cellcolor{blue!10}\textbf{63.18}$_{\pm 0.13}$ \\
& Pass@8
& 84.42$_{\pm 0.20}$
& 86.03$_{\pm 0.48}$
& 85.95$_{\pm 0.34}$
& 85.18$_{\pm 0.15}$
& \cellcolor{blue!10}\textbf{87.20}$_{\pm 0.39}$
& \cellcolor{blue!10}\underline{86.92}$_{\pm 0.38}$ \\
& Pass@64
& 94.40$_{\pm 0.20}$
& \underline{95.13}$_{\pm 0.50}$
& 95.00$_{\pm 0.20}$
& 94.73$_{\pm 0.46}$
& \cellcolor{blue!10}\textbf{95.67}$_{\pm 0.50}$
& \cellcolor{blue!10}\underline{95.13}$_{\pm 0.81}$ \\

\midrule
\multirow{3}{*}{MATH500}
& Avg@64
& 46.85$_{\pm 0.20}$
& 50.08$_{\pm 0.35}$
& 50.13$_{\pm 0.08}$
& 49.53$_{\pm 0.11}$
& \cellcolor{blue!10}\underline{51.71}$_{\pm 0.48}$
& \cellcolor{blue!10}\textbf{51.79}$_{\pm 0.69}$ \\
& Pass@8
& 75.38$_{\pm 0.07}$
& 78.91$_{\pm 0.40}$
& 79.02$_{\pm 0.09}$
& 78.52$_{\pm 0.36}$
& \cellcolor{blue!10}\textbf{80.08}$_{\pm 0.31}$
& \cellcolor{blue!10}\underline{79.75}$_{\pm 0.65}$ \\
& Pass@64
& 90.20$_{\pm 0.40}$
& \textbf{93.07}$_{\pm 0.42}$
& 92.47$_{\pm 0.23}$
& \underline{92.93}$_{\pm 0.81}$
& \cellcolor{blue!10}\underline{92.93}$_{\pm 0.83}$
& \cellcolor{blue!10}92.53$_{\pm 0.61}$ \\

\midrule
\multirow{3}{*}{Minerva}
& Avg@64
& 11.95$_{\pm 0.24}$
& \textbf{13.64}$_{\pm 0.23}$
& \underline{13.57}$_{\pm 0.21}$
& 12.94$_{\pm 0.16}$
& \cellcolor{blue!10}13.13$_{\pm 0.19}$
& \cellcolor{blue!10}13.37$_{\pm 0.20}$ \\
& Pass@8
& 28.47$_{\pm 0.72}$
& \underline{31.52}$_{\pm 0.31}$
& \textbf{31.57}$_{\pm 0.23}$
& 30.83$_{\pm 0.38}$
& \cellcolor{blue!10}31.08$_{\pm 0.67}$
& \cellcolor{blue!10}31.04$_{\pm 0.33}$ \\
& Pass@64
& 41.30$_{\pm 1.70}$
& \underline{45.46}$_{\pm 0.76}$
& \textbf{45.71}$_{\pm 1.18}$
& 44.98$_{\pm 0.93}$
& \cellcolor{blue!10}44.12$_{\pm 1.84}$
& \cellcolor{blue!10}45.10$_{\pm 2.78}$ \\

\midrule
\multirow{3}{*}{\textbf{Average}}
& Avg@64
& 24.27$_{\pm 0.15}$
& 26.40$_{\pm 0.16}$
& 26.44$_{\pm 0.12}$
& 25.77$_{\pm 0.12}$
& \cellcolor{blue!10}\underline{27.11}$_{\pm 0.19}$
& \cellcolor{blue!10}\textbf{27.22}$_{\pm 0.19}$ \\
& Pass@8
& 43.28$_{\pm 0.29}$
& 47.05$_{\pm 0.49}$
& 47.20$_{\pm 0.32}$
& 46.77$_{\pm 0.34}$
& \cellcolor{blue!10}\textbf{47.39}$_{\pm 0.22}$
& \cellcolor{blue!10}\underline{47.29}$_{\pm 0.45}$ \\
& Pass@64
& 58.16$_{\pm 0.99}$
& \underline{62.65}$_{\pm 1.12}$
& \textbf{62.89}$_{\pm 1.26}$
& 61.88$_{\pm 1.83}$
& \cellcolor{blue!10}60.59$_{\pm 0.62}$
& \cellcolor{blue!10}62.13$_{\pm 2.18}$ \\

\midrule
\multicolumn{8}{l}{\textbf{\textit{General capabilities}}} \\
\midrule

HellaSwag
& Avg@8
& 43.37$_{\pm 0.18}$
& 42.18$_{\pm 0.12}$
& 42.18$_{\pm 0.05}$
& 43.22$_{\pm 0.04}$
& \cellcolor{blue!10}\underline{44.66}$_{\pm 0.27}$
& \cellcolor{blue!10}\textbf{44.72}$_{\pm 0.17}$ \\

IFEval
& Avg@8
& \textbf{41.03}$_{\pm 1.44}$
& 35.37$_{\pm 1.02}$
& 33.95$_{\pm 1.68}$
& \underline{39.18}$_{\pm 1.15}$
& \cellcolor{blue!10}35.37$_{\pm 1.02}$
& \cellcolor{blue!10}38.39$_{\pm 1.36}$ \\

MMLU
& Avg@8
& 45.46$_{\pm 0.23}$
& 45.07$_{\pm 0.14}$
& \underline{45.75}$_{\pm 0.37}$
& 45.53$_{\pm 0.19}$
& \cellcolor{blue!10}\textbf{46.19}$_{\pm 0.43}$
& \cellcolor{blue!10}45.72$_{\pm 0.25}$ \\

\bottomrule
\end{tabular}%
}
\end{table*}

\begin{table*}[h]
\centering
\caption{Online distillation results on Qwen3-0.6B-Base using MATH. Results are reported as mean$_{\pm \text{std}}$ across three seeds. \textbf{Bold} marks the best result and \underline{underlining} marks the second-best result; ties receive the same formatting.}
\label{tab:qwen3_06b_base_math_kd}
\resizebox{\textwidth}{!}{%
\begin{tabular}{ll ccccccc}
\toprule
\multirow{2}{*}{\textbf{Benchmark}}
& \multirow{2}{*}{\textbf{Metric}}
& \multicolumn{7}{c}{\textbf{Qwen3-0.6B-Base} (MATH)} \\
\cmidrule(lr){3-9}
&
& \textbf{FKL}
& \textbf{OPD}
& \textbf{EOPD}
& \textbf{Skew-KL}
& \textbf{TAID}
& \cellcolor{blue!10}\textbf{ConMix-FKL}
& \cellcolor{blue!10}\textbf{GeoMix-FKL} \\
\midrule
\multicolumn{9}{l}{\textbf{\textit{Mathematical reasoning}}} \\
\midrule

\multirow{3}{*}{AIME24}
& Avg@64
& 1.15$_{\pm 0.29}$
& 0.88$_{\pm 0.27}$
& 1.18$_{\pm 0.13}$
& \underline{1.39}$_{\pm 0.29}$
& 1.08$_{\pm 0.22}$
& \cellcolor{blue!10}\textbf{1.58}$_{\pm 0.13}$
& \cellcolor{blue!10}1.37$_{\pm 0.18}$ \\
& Pass@8
& 6.44$_{\pm 1.33}$
& 4.66$_{\pm 1.00}$
& 4.95$_{\pm 1.27}$
& \underline{6.62}$_{\pm 0.83}$
& 4.92$_{\pm 0.47}$
& \cellcolor{blue!10}\textbf{6.97}$_{\pm 0.69}$
& \cellcolor{blue!10}6.02$_{\pm 0.11}$ \\
& Pass@64
& \underline{22.22}$_{\pm 5.09}$
& 17.78$_{\pm 3.85}$
& 12.22$_{\pm 5.09}$
& \underline{22.22}$_{\pm 1.92}$
& 16.67$_{\pm 3.34}$
& \cellcolor{blue!10}\textbf{23.33}$_{\pm 5.77}$
& \cellcolor{blue!10}21.11$_{\pm 0.00}$ \\

\midrule
\multirow{3}{*}{AIME25}
& Avg@64
& 0.47$_{\pm 0.05}$
& 0.56$_{\pm 0.16}$
& \textbf{0.83}$_{\pm 0.14}$
& 0.50$_{\pm 0.26}$
& 0.47$_{\pm 0.11}$
& \cellcolor{blue!10}\underline{0.61}$_{\pm 0.32}$
& \cellcolor{blue!10}\textbf{0.83}$_{\pm 0.23}$ \\
& Pass@8
& 3.47$_{\pm 0.50}$
& 3.97$_{\pm 1.33}$
& \textbf{5.72}$_{\pm 0.95}$
& 3.69$_{\pm 1.84}$
& 3.57$_{\pm 0.81}$
& \cellcolor{blue!10}4.30$_{\pm 2.09}$
& \cellcolor{blue!10}\underline{5.54}$_{\pm 1.44}$ \\
& Pass@64
& 15.56$_{\pm 5.09}$
& \underline{18.89}$_{\pm 10.18}$
& \textbf{20.00}$_{\pm 5.77}$
& 15.55$_{\pm 6.94}$
& \underline{18.89}$_{\pm 3.85}$
& \cellcolor{blue!10}16.66$_{\pm 5.77}$
& \cellcolor{blue!10}\textbf{20.00}$_{\pm 3.85}$ \\

\midrule
\multirow{3}{*}{AMC23}
& Avg@64
& 17.75$_{\pm 0.32}$
& 15.28$_{\pm 0.67}$
& 19.78$_{\pm 0.55}$
& 16.93$_{\pm 0.32}$
& 18.31$_{\pm 0.22}$
& \cellcolor{blue!10}\underline{20.04}$_{\pm 0.35}$
& \cellcolor{blue!10}\textbf{21.12}$_{\pm 0.47}$ \\
& Pass@8
& 48.47$_{\pm 1.54}$
& 47.57$_{\pm 0.59}$
& 49.62$_{\pm 2.68}$
& 49.20$_{\pm 1.46}$
& 48.80$_{\pm 0.41}$
& \cellcolor{blue!10}\underline{50.95}$_{\pm 1.63}$
& \cellcolor{blue!10}\textbf{51.65}$_{\pm 1.35}$ \\
& Pass@64
& 80.00$_{\pm 5.00}$
& 76.67$_{\pm 2.89}$
& 74.17$_{\pm 8.04}$
& 80.00$_{\pm 2.50}$
& 78.33$_{\pm 3.82}$
& \cellcolor{blue!10}\underline{80.83}$_{\pm 8.04}$
& \cellcolor{blue!10}\textbf{81.67}$_{\pm 6.29}$ \\

\midrule
\multirow{3}{*}{GSM8K}
& Avg@64
& 61.24$_{\pm 0.04}$
& 47.26$_{\pm 1.80}$
& 62.11$_{\pm 0.25}$
& 61.52$_{\pm 0.12}$
& 62.90$_{\pm 0.09}$
& \cellcolor{blue!10}\underline{63.16}$_{\pm 0.34}$
& \cellcolor{blue!10}\textbf{64.80}$_{\pm 0.20}$ \\
& Pass@8
& 88.38$_{\pm 0.24}$
& 85.47$_{\pm 0.77}$
& 88.14$_{\pm 0.28}$
& 88.63$_{\pm 0.20}$
& \textbf{88.95}$_{\pm 0.10}$
& \cellcolor{blue!10}88.65$_{\pm 0.47}$
& \cellcolor{blue!10}\underline{88.81}$_{\pm 0.03}$ \\
& Pass@64
& \textbf{96.60}$_{\pm 0.35}$
& 96.07$_{\pm 0.81}$
& 96.33$_{\pm 0.23}$
& 96.07$_{\pm 0.81}$
& \underline{96.47}$_{\pm 0.42}$
& \cellcolor{blue!10}96.40$_{\pm 0.72}$
& \cellcolor{blue!10}\textbf{96.60}$_{\pm 0.35}$ \\

\midrule
\multirow{3}{*}{MATH500}
& Avg@64
& 40.98$_{\pm 0.09}$
& 35.21$_{\pm 1.00}$
& 43.63$_{\pm 0.16}$
& 40.27$_{\pm 0.16}$
& 42.46$_{\pm 0.10}$
& \cellcolor{blue!10}\underline{43.87}$_{\pm 0.14}$
& \cellcolor{blue!10}\textbf{45.83}$_{\pm 0.21}$ \\
& Pass@8
& 70.87$_{\pm 0.66}$
& 69.63$_{\pm 0.88}$
& 72.48$_{\pm 0.38}$
& 70.90$_{\pm 0.10}$
& 72.16$_{\pm 0.19}$
& \cellcolor{blue!10}\underline{72.63}$_{\pm 0.78}$
& \cellcolor{blue!10}\textbf{73.69}$_{\pm 0.34}$ \\
& Pass@64
& 88.20$_{\pm 0.72}$
& 87.87$_{\pm 0.76}$
& 88.73$_{\pm 0.50}$
& 88.33$_{\pm 0.31}$
& \underline{88.87}$_{\pm 0.46}$
& \cellcolor{blue!10}\underline{88.87}$_{\pm 0.83}$
& \cellcolor{blue!10}\textbf{89.60}$_{\pm 0.40}$ \\

\midrule
\multirow{3}{*}{Minerva}
& Avg@64
& 11.82$_{\pm 0.06}$
& 10.07$_{\pm 0.16}$
& 12.49$_{\pm 0.09}$
& 12.28$_{\pm 0.32}$
& 12.49$_{\pm 0.25}$
& \cellcolor{blue!10}\underline{12.72}$_{\pm 0.12}$
& \cellcolor{blue!10}\textbf{13.21}$_{\pm 0.12}$ \\
& Pass@8
& 30.65$_{\pm 0.16}$
& 29.27$_{\pm 0.24}$
& 31.24$_{\pm 0.48}$
& 31.23$_{\pm 0.40}$
& 31.28$_{\pm 0.49}$
& \cellcolor{blue!10}\underline{31.64}$_{\pm 0.15}$
& \cellcolor{blue!10}\textbf{31.91}$_{\pm 0.22}$ \\
& Pass@64
& 45.96$_{\pm 1.68}$
& 45.59$_{\pm 1.94}$
& 46.94$_{\pm 1.53}$
& 46.57$_{\pm 0.77}$
& 46.69$_{\pm 2.21}$
& \cellcolor{blue!10}\underline{47.43}$_{\pm 1.27}$
& \cellcolor{blue!10}\textbf{47.67}$_{\pm 1.88}$ \\

\midrule
\multirow{3}{*}{\textbf{Average}}
& Avg@64
& 22.23$_{\pm 0.14}$
& 18.21$_{\pm 0.68}$
& 23.34$_{\pm 0.22}$
& 22.15$_{\pm 0.25}$
& 22.95$_{\pm 0.17}$
& \cellcolor{blue!10}\underline{23.66}$_{\pm 0.23}$
& \cellcolor{blue!10}\textbf{24.53}$_{\pm 0.23}$ \\
& Pass@8
& 41.38$_{\pm 0.74}$
& 40.09$_{\pm 0.80}$
& 42.02$_{\pm 1.01}$
& 41.71$_{\pm 0.81}$
& 41.61$_{\pm 0.41}$
& \cellcolor{blue!10}\underline{42.52}$_{\pm 0.97}$
& \cellcolor{blue!10}\textbf{42.94}$_{\pm 0.58}$ \\
& Pass@64
& 58.09$_{\pm 2.99}$
& 57.15$_{\pm 3.40}$
& 56.40$_{\pm 3.53}$
& 58.12$_{\pm 2.21}$
& 57.65$_{\pm 2.35}$
& \cellcolor{blue!10}\underline{58.92}$_{\pm 3.73}$
& \cellcolor{blue!10}\textbf{59.44}$_{\pm 2.13}$ \\

\midrule
\multicolumn{9}{l}{\textbf{\textit{General capabilities}}} \\
\midrule

HellaSwag
& Avg@8
& 48.67$_{\pm 0.08}$
& 48.27$_{\pm 0.11}$
& 48.60$_{\pm 0.05}$
& 48.55$_{\pm 0.01}$
& \textbf{48.73}$_{\pm 0.08}$
& \cellcolor{blue!10}\underline{48.71}$_{\pm 0.05}$
& \cellcolor{blue!10}48.61$_{\pm 0.06}$ \\

IFEval
& Avg@8
& 20.52$_{\pm 0.55}$
& 20.27$_{\pm 0.59}$
& 20.46$_{\pm 1.30}$
& 20.58$_{\pm 1.20}$
& \underline{21.44}$_{\pm 1.33}$
& \cellcolor{blue!10}\textbf{22.43}$_{\pm 1.34}$
& \cellcolor{blue!10}21.20$_{\pm 0.28}$ \\

MMLU
& Avg@8
& 37.87$_{\pm 0.53}$
& 29.27$_{\pm 0.22}$
& 35.87$_{\pm 0.37}$
& 37.39$_{\pm 0.29}$
& 38.10$_{\pm 0.37}$
& \cellcolor{blue!10}\underline{38.60}$_{\pm 0.27}$
& \cellcolor{blue!10}\textbf{38.87}$_{\pm 0.79}$ \\

\bottomrule
\end{tabular}%
}
\end{table*}

\FloatBarrier
\subsection{Code Generation} \label{app:instruction_following}
We additionally test Qwen3-1.7B on a coding task using LeetCodeDataset~\citep{xia2025leetcodedatasettemporaldatasetrobust}. The dataset contains 2,869 Python LeetCode problems with broad algorithmic coverage and approximately 100 unit tests per problem. Training prompts are directly taken from the \texttt{query} column (see chat template below). This setting follows the same offline and online distillation protocols as the main experiments, but replaces mathematical reasoning with code generation.

\begin{center}
\begin{tcolorbox}[colback=gray!5!white, colframe=gray!50!black, width=0.6\textwidth, sharp corners]
\begin{verbatim}
<|im_start|>user
{query}<|im_end|>
<|im_start|>assistant
<think>

</think>
\end{verbatim}
\end{tcolorbox}
\end{center}

For evaluation, each generated solution is executed in a sandboxed verifier adapted from the \texttt{prime\_code} verifier in \texttt{verl}. We include problem-specific imports and data structures, such as \texttt{ListNode} and \texttt{TreeNode}. The reward is computed as the fraction of successful unit tests. We report Avg@64, the average unit-test pass rate over 64 samples, and Pass@k, the probability that at least one of $k$ samples passes all unit tests.

Tables~\ref{tab:qwen3_code}-\ref{tab:qwen3_base_code} show that entropy-aware mixing transfers to coding in both regimes: (i) offline, ConMix-SFT achieves the best Avg@64 and Pass@1, while GeoMix-SFT remains competitive; (ii) online, GeoMix-FKL gives the strongest results, with the best Avg@64, Pass@1, and Pass@8, and a tie for best Pass@64. This suggests that entropy-based guidance extends beyond mathematical reasoning to executable code generation.

\begin{table}[h]
\centering
\caption{Offline distillation results on Qwen3-1.7B using LeetCodeDataset. Results are reported as mean$_{\pm\text{std}}$ over three seeds. \textbf{Bold} indicates the best result and \underline{underlining} the second-best result within each setting, with ties formatted identically.}
\label{tab:qwen3_code}
\resizebox{\columnwidth}{!}{%
\begin{tabular}{llcccccc}
\toprule
\multirow{2}{*}{\textbf{Benchmark}} &
\multirow{2}{*}{\textbf{Metric}} &
\multicolumn{6}{c}{\textbf{Qwen3-1.7B} (LeetCodeDataset)} \\
\cmidrule(lr){3-8}
&
& \textbf{\makecell{Student-\\generated}}
& \textbf{\makecell{Teacher-\\generated}}
& \textbf{SKD}
& \textbf{EAFT}
& \cellcolor{blue!10}\textbf{ConMix-SFT}
& \cellcolor{blue!10}\textbf{GeoMix-SFT} \\
\midrule

\multicolumn{8}{l}{\textbf{\textit{Coding performance}}} \\
\midrule

\multirow{4}{*}{LeetCode}
& Avg@64
& 25.95$_{\pm0.77}$ & 24.64$_{\pm0.66}$ & 1.32$_{\pm0.27}$ & 24.54$_{\pm0.78}$
& \cellcolor{blue!10}\textbf{26.63}$_{\pm0.95}$
& \cellcolor{blue!10}\underline{26.29}$_{\pm0.29}$ \\

& Pass@1
& 11.55$_{\pm0.03}$ & 10.74$_{\pm0.42}$ & 7.57$_{\pm1.56}$ & 10.45$_{\pm0.46}$
& \cellcolor{blue!10}\textbf{12.24}$_{\pm0.29}$
& \cellcolor{blue!10}\underline{12.13}$_{\pm0.06}$ \\

& Pass@8
& 18.50$_{\pm1.40}$ & 22.73$_{\pm0.42}$ & 7.57$_{\pm1.56}$ & 22.62$_{\pm0.45}$
& \cellcolor{blue!10}\textbf{22.90}$_{\pm0.77}$
& \cellcolor{blue!10}\underline{22.83}$_{\pm0.69}$ \\

& Pass@64
& 22.68$_{\pm2.81}$ & \textbf{33.92}$_{\pm1.16}$ & 21.11$_{\pm5.09}$
& \underline{33.19}$_{\pm0.67}$
& \cellcolor{blue!10}32.16$_{\pm1.24}$
& \cellcolor{blue!10}32.75$_{\pm0.67}$ \\




\bottomrule
\end{tabular}
}
\end{table}

\begin{table}[h]
\centering
\caption{Online distillation results on Qwen3-1.7B-Base using LeetCodeDataset. Results are reported as mean$_{\pm\text{std}}$ over three seeds. \textbf{Bold} indicates the best result and \underline{underlining} the second-best result within each setting, with ties formatted identically.}
\label{tab:qwen3_base_code}
\resizebox{\columnwidth}{!}{%
\begin{tabular}{llccccccc}
\toprule
\multirow{2}{*}{\textbf{Benchmark}} &
\multirow{2}{*}{\textbf{Metric}} &
\multicolumn{7}{c}{\textbf{Qwen3-1.7B-Base} (LeetCodeDataset)} \\
\cmidrule(lr){3-9}
&
& \textbf{FKL}
& \textbf{OPD}
& \textbf{EOPD}
& \textbf{Skew-KL}
& \textbf{TAID}
& \cellcolor{blue!10}\textbf{ConMix-FKL}
& \cellcolor{blue!10}\textbf{GeoMix-FKL} \\
\midrule

\multicolumn{9}{l}{\textbf{\textit{Coding performance}}} \\
\midrule

\multirow{4}{*}{LeetCode}
& Avg@64
& 19.86$_{\pm0.12}$ & 5.27$_{\pm0.09}$ & 7.30$_{\pm0.53}$
& 20.14$_{\pm0.09}$ & \underline{20.33}$_{\pm0.29}$
& \cellcolor{blue!10}20.14$_{\pm0.06}$
& \cellcolor{blue!10}\textbf{20.35}$_{\pm0.25}$ \\

& Pass@1
& 6.03$_{\pm0.03}$ & 1.67$_{\pm0.05}$ & 2.37$_{\pm0.10}$
& 6.11$_{\pm0.09}$ & \underline{6.30}$_{\pm0.12}$
& \cellcolor{blue!10}6.11$_{\pm0.09}$
& \cellcolor{blue!10}\textbf{6.37}$_{\pm0.12}$ \\

& Pass@8
& 18.51$_{\pm0.14}$ & 9.67$_{\pm0.21}$ & 11.90$_{\pm0.37}$
& 17.97$_{\pm0.25}$ & \underline{18.76}$_{\pm0.59}$
& \cellcolor{blue!10}17.97$_{\pm0.18}$
& \cellcolor{blue!10}\textbf{18.92}$_{\pm0.38}$ \\

& Pass@64
& \textbf{29.96}$_{\pm0.44}$ & 24.08$_{\pm0.25}$ & 24.52$_{\pm0.67}$
& 27.60$_{\pm0.51}$ & \textbf{29.96}$_{\pm1.17}$
& \cellcolor{blue!10}\underline{27.90}$_{\pm1.34}$
& \cellcolor{blue!10}\textbf{29.96}$_{\pm1.59}$ \\




\bottomrule
\end{tabular}
}
\end{table}

\FloatBarrier
\section{Results: Alternative Mixture Distributions} \label{app:results_amid}
We further evaluate Entropy-Aware Mixing across the generalised AMiD mixture family. In addition to the convex and geometric mixtures considered in the main paper, we test AMiD settings with $\beta \in \{-5,-3,0\}$; see Appendix~\ref{app:amid} for the generalised mixture definition. We follow the experimental setup in Section~\ref{sec:experiment}, with hyperparameters detailed in Appendix~\ref{app:hyper_params}, and conduct all experiments in the online distillation setting.

For each $\beta$, Table~\ref{tab:full_qwen3_base_amid} compares a fixed interpolation coefficient ($\alpha_t=0.1$) against our entropy-aware coefficient ($\alpha_t=f(\tilde{H}_t)$). Across the evaluated AMiD mixtures, entropy-aware interpolation improves mathematical reasoning performance over the corresponding fixed-coefficient baseline. These results show that the proposed entropy-based gating strategy is not specific to ConMix or GeoMix, but can also be combined effectively with other mixture forms.

\begin{table*}[h]
\centering
\caption{AMiD results on Qwen3-1.7B-Base using MATH. Results are reported as mean$_{\pm \text{std}}$ across three seeds. Within each AMiD $\beta$ setting, \textbf{bold} marks the better result and \underline{underlining} marks the second-best result; ties receive the same formatting.}
\label{tab:full_qwen3_base_amid}
\resizebox{\textwidth}{!}{%
\begin{tabular}{ll cc cc cc}
\toprule
\multirow{2}{*}{\textbf{Benchmark}}
& \multirow{2}{*}{\textbf{Metric}}
& \multicolumn{2}{c}{AMiD $\beta=-5$}
& \multicolumn{2}{c}{AMiD $\beta=-3$}
& \multicolumn{2}{c}{AMiD $\beta=0$} \\
\cmidrule(lr){3-4}
\cmidrule(lr){5-6}
\cmidrule(lr){7-8}
& & \textbf{$\alpha_t=0.1$} & \cellcolor{blue!10}\textbf{$\alpha_t = f(\tilde{H}_t)$}
  & \textbf{$\alpha_t=0.1$} & \cellcolor{blue!10}\textbf{$\alpha_t = f(\tilde{H}_t)$}
  & \textbf{$\alpha_t=0.1$} & \cellcolor{blue!10}\textbf{$\alpha_t = f(\tilde{H}_t)$} \\
\midrule

\multicolumn{8}{l}{\textbf{\textit{Mathematical reasoning}}} \\
\midrule

\multirow{3}{*}{AIME24}
& Avg@64
& $\mathbf{5.33}_{\pm 0.13}$ & \cellcolor{blue!10}$\underline{5.23}_{\pm 0.03}$
& $\underline{4.31}_{\pm 0.42}$ & \cellcolor{blue!10}$\mathbf{5.72}_{\pm 0.27}$
& $\underline{5.92}_{\pm 0.20}$ & \cellcolor{blue!10}$\mathbf{7.22}_{\pm 0.35}$ \\
& Pass@8
& $\mathbf{18.95}_{\pm 0.64}$ & \cellcolor{blue!10}$\underline{17.07}_{\pm 0.45}$
& $\underline{15.41}_{\pm 0.71}$ & \cellcolor{blue!10}$\mathbf{18.33}_{\pm 1.85}$
& $\underline{20.18}_{\pm 1.24}$ & \cellcolor{blue!10}$\mathbf{21.40}_{\pm 0.76}$ \\
& Pass@64
& $\mathbf{35.56}_{\pm 1.92}$ & \cellcolor{blue!10}$\underline{32.78}_{\pm 0.96}$
& $\mathbf{35.56}_{\pm 1.92}$ & \cellcolor{blue!10}$\underline{32.22}_{\pm 5.09}$
& $\mathbf{37.22}_{\pm 5.36}$ & \cellcolor{blue!10}$\underline{36.67}_{\pm 6.67}$ \\
\midrule

\multirow{3}{*}{AIME25}
& Avg@64
& $\underline{3.04}_{\pm 0.03}$ & \cellcolor{blue!10}$\mathbf{3.65}_{\pm 0.24}$
& $\underline{3.26}_{\pm 0.18}$ & \cellcolor{blue!10}$\mathbf{3.80}_{\pm 0.36}$
& $\underline{4.01}_{\pm 0.11}$ & \cellcolor{blue!10}$\mathbf{4.64}_{\pm 0.07}$ \\
& Pass@8
& $\underline{13.30}_{\pm 0.48}$ & \cellcolor{blue!10}$\mathbf{14.51}_{\pm 1.01}$
& $\underline{14.67}_{\pm 1.07}$ & \cellcolor{blue!10}$\mathbf{15.10}_{\pm 1.06}$
& $\underline{14.51}_{\pm 1.59}$ & \cellcolor{blue!10}$\mathbf{15.48}_{\pm 0.70}$ \\
& Pass@64
& $\underline{31.11}_{\pm 3.85}$ & \cellcolor{blue!10}$\mathbf{33.89}_{\pm 2.55}$
& $\underline{33.89}_{\pm 7.52}$ & \cellcolor{blue!10}$\mathbf{36.11}_{\pm 2.55}$
& $\mathbf{34.44}_{\pm 1.92}$ & \cellcolor{blue!10}$\underline{32.22}_{\pm 7.70}$ \\
\midrule

\multirow{3}{*}{AMC23}
& Avg@64
& $\underline{29.31}_{\pm 0.95}$ & \cellcolor{blue!10}$\mathbf{30.92}_{\pm 0.61}$
& $\underline{30.56}_{\pm 0.28}$ & \cellcolor{blue!10}$\mathbf{31.88}_{\pm 0.15}$
& $\underline{32.93}_{\pm 0.22}$ & \cellcolor{blue!10}$\mathbf{35.87}_{\pm 0.41}$ \\
& Pass@8
& $\underline{65.53}_{\pm 0.68}$ & \cellcolor{blue!10}$\mathbf{66.23}_{\pm 0.57}$
& $\underline{66.41}_{\pm 0.14}$ & \cellcolor{blue!10}$\mathbf{66.53}_{\pm 1.00}$
& $\underline{68.44}_{\pm 0.27}$ & \cellcolor{blue!10}$\mathbf{69.59}_{\pm 1.08}$ \\
& Pass@64
& $\mathbf{90.00}_{\pm 2.17}$ & \cellcolor{blue!10}$\underline{86.67}_{\pm 1.44}$
& $\underline{88.75}_{\pm 3.31}$ & \cellcolor{blue!10}$\mathbf{90.42}_{\pm 4.39}$
& $\mathbf{90.00}_{\pm 2.50}$ & \cellcolor{blue!10}$\underline{89.58}_{\pm 5.05}$ \\
\midrule

\multirow{3}{*}{GSM8K}
& Avg@64
& $\underline{78.73}_{\pm 0.18}$ & \cellcolor{blue!10}$\mathbf{79.66}_{\pm 0.12}$
& $\underline{78.74}_{\pm 0.08}$ & \cellcolor{blue!10}$\mathbf{79.87}_{\pm 0.17}$
& $\underline{79.53}_{\pm 0.13}$ & \cellcolor{blue!10}$\mathbf{80.81}_{\pm 0.17}$ \\
& Pass@8
& $\mathbf{95.40}_{\pm 0.27}$ & \cellcolor{blue!10}$\underline{95.25}_{\pm 0.09}$
& $\underline{95.27}_{\pm 0.04}$ & \cellcolor{blue!10}$\mathbf{95.59}_{\pm 0.06}$
& $\underline{95.52}_{\pm 0.05}$ & \cellcolor{blue!10}$\mathbf{95.73}_{\pm 0.20}$ \\
& Pass@64
& $\mathbf{99.07}_{\pm 0.29}$ & \cellcolor{blue!10}$\underline{99.00}_{\pm 0.20}$
& $\underline{98.97}_{\pm 0.25}$ & \cellcolor{blue!10}$\mathbf{99.33}_{\pm 0.12}$
& $\mathbf{99.53}_{\pm 0.31}$ & \cellcolor{blue!10}$\underline{99.33}_{\pm 0.12}$ \\
\midrule

\multirow{3}{*}{MATH500}
& Avg@64
& $\underline{57.69}_{\pm 0.17}$ & \cellcolor{blue!10}$\mathbf{59.32}_{\pm 0.14}$
& $\underline{57.58}_{\pm 0.20}$ & \cellcolor{blue!10}$\mathbf{60.09}_{\pm 0.13}$
& $\underline{58.38}_{\pm 0.10}$ & \cellcolor{blue!10}$\mathbf{62.90}_{\pm 0.02}$ \\
& Pass@8
& $\underline{83.83}_{\pm 0.35}$ & \cellcolor{blue!10}$\mathbf{84.20}_{\pm 0.25}$
& $\underline{83.43}_{\pm 0.21}$ & \cellcolor{blue!10}$\mathbf{84.91}_{\pm 0.17}$
& $\underline{84.54}_{\pm 0.09}$ & \cellcolor{blue!10}$\mathbf{85.83}_{\pm 0.24}$ \\
& Pass@64
& $\mathbf{93.07}_{\pm 0.60}$ & \cellcolor{blue!10}$\underline{92.87}_{\pm 0.84}$
& $\underline{93.13}_{\pm 0.12}$ & \cellcolor{blue!10}$\mathbf{93.80}_{\pm 0.89}$
& $\underline{93.53}_{\pm 0.23}$ & \cellcolor{blue!10}$\mathbf{94.07}_{\pm 0.23}$ \\
\midrule

\multirow{3}{*}{Minerva}
& Avg@64
& $\underline{22.75}_{\pm 0.01}$ & \cellcolor{blue!10}$\mathbf{23.54}_{\pm 0.20}$
& $\underline{22.79}_{\pm 0.08}$ & \cellcolor{blue!10}$\mathbf{23.55}_{\pm 0.14}$
& $\underline{23.49}_{\pm 0.10}$ & \cellcolor{blue!10}$\mathbf{24.25}_{\pm 0.17}$ \\
& Pass@8
& $\underline{44.30}_{\pm 0.50}$ & \cellcolor{blue!10}$\mathbf{44.31}_{\pm 0.32}$
& $\underline{44.26}_{\pm 0.51}$ & \cellcolor{blue!10}$\mathbf{44.58}_{\pm 0.49}$
& $\underline{44.15}_{\pm 0.37}$ & \cellcolor{blue!10}$\mathbf{44.54}_{\pm 0.53}$ \\
& Pass@64
& $\mathbf{57.97}_{\pm 1.18}$ & \cellcolor{blue!10}$\underline{57.78}_{\pm 1.61}$
& $\underline{56.86}_{\pm 1.38}$ & \cellcolor{blue!10}$\mathbf{58.33}_{\pm 0.42}$
& $\mathbf{57.60}_{\pm 1.22}$ & \cellcolor{blue!10}$\underline{56.74}_{\pm 0.91}$ \\
\midrule

\multirow{3}{*}{\textbf{Average}}
& Avg@64
& $\underline{32.81}_{\pm 0.20}$ & \cellcolor{blue!10}$\mathbf{33.72}_{\pm 0.10}$
& $\underline{32.87}_{\pm 0.03}$ & \cellcolor{blue!10}$\mathbf{34.15}_{\pm 0.08}$
& $\underline{34.04}_{\pm 0.04}$ & \cellcolor{blue!10}$\mathbf{35.95}_{\pm 0.05}$ \\
& Pass@8
& $\underline{53.55}_{\pm 0.02}$ & \cellcolor{blue!10}$\mathbf{53.60}_{\pm 0.23}$
& $\underline{53.24}_{\pm 0.37}$ & \cellcolor{blue!10}$\mathbf{54.17}_{\pm 0.16}$
& $\underline{54.56}_{\pm 0.13}$ & \cellcolor{blue!10}$\mathbf{55.43}_{\pm 0.15}$ \\
& Pass@64
& $\mathbf{67.79}_{\pm 0.52}$ & \cellcolor{blue!10}$\underline{67.16}_{\pm 0.66}$
& $\underline{67.86}_{\pm 2.06}$ & \cellcolor{blue!10}$\mathbf{68.37}_{\pm 1.91}$
& $\mathbf{68.72}_{\pm 1.27}$ & \cellcolor{blue!10}$\underline{68.10}_{\pm 1.95}$ \\





\bottomrule
\end{tabular}%
}
\end{table*}

\FloatBarrier
\section{Results: Reverse-KL Distillation with OPD} \label{app:opd}
Our main online experiments use forward KL to train the student directly on the proposed mixed target distribution. We also extend Entropy-Aware Mixing to OPD~\citep{lu2025onpolicydistillation} by replacing its teacher target with the geometric mixture, yielding GeoMix-OPD. In our implementation, OPD evaluates the teacher probability assigned to the token sampled by the student. Since the truncated top-$K$ teacher probabilities are used to construct the mixture, we add the sampled token to the evaluated support when necessary. This requires additional compute because the corresponding probability must be computed under the mixed distribution, which may require renormalisation.

Table~\ref{tab:opd} compares standard OPD with GeoMix-OPD using the same Qwen3-1.7B-Base setup as in our main online experiments. GeoMix-OPD remains competitive with OPD across benchmarks, with an average +0.08 Avg@64, +0.04 Pass@8, and +1.40 Pass@64 improvement. Our adaptive mixtures are not limited to the forward-KL objective studied in our main experiments and can also be applied to other divergences. As noted earlier, we leave a systematic study of alternative divergence objectives to future work.

\begin{table*}[h]
\centering
\caption{Comparison between OPD and GeoMix-OPD on mathematical reasoning benchmarks. Results are reported as mean$_{\pm \text{std}}$ across three seeds. \textbf{Bold} marks the best result and \underline{underlining} marks the second-best result; ties receive the same formatting.}
\label{tab:opd}
\begin{tabular}{ll cc}
\toprule
\multirow{2}{*}{\textbf{Benchmark}}
& \multirow{2}{*}{\textbf{Metric}}
& \multicolumn{2}{c}{\textbf{Qwen3-1.7B-Base} (MATH)} \\
\cmidrule(lr){3-4}
&
& \textbf{OPD}
& \cellcolor{blue!10}\textbf{GeoMix-OPD} \\
\midrule
\multicolumn{4}{l}{\textbf{\textit{Mathematical reasoning}}} \\
\midrule

\multirow{3}{*}{AIME24}
& Avg@64
& \underline{7.63}$_{\pm 0.26}$
& \cellcolor{blue!10}\textbf{7.86}$_{\pm 0.27}$ \\
& Pass@8
& \underline{21.15}$_{\pm 0.85}$
& \cellcolor{blue!10}\textbf{22.23}$_{\pm 0.95}$ \\
& Pass@64
& \underline{35.57}$_{\pm 1.77}$
& \cellcolor{blue!10}\textbf{41.50}$_{\pm 2.87}$ \\

\midrule
\multirow{3}{*}{AIME25}
& Avg@64
& \textbf{4.87}$_{\pm 0.10}$
& \cellcolor{blue!10}\underline{4.28}$_{\pm 0.15}$ \\
& Pass@8
& \textbf{16.27}$_{\pm 0.44}$
& \cellcolor{blue!10}\underline{14.85}$_{\pm 0.46}$ \\
& Pass@64
& \underline{33.06}$_{\pm 1.60}$
& \cellcolor{blue!10}\textbf{35.27}$_{\pm 1.68}$ \\

\midrule
\multirow{3}{*}{AMC23}
& Avg@64
& \underline{37.00}$_{\pm 0.43}$
& \cellcolor{blue!10}\textbf{37.81}$_{\pm 0.08}$ \\
& Pass@8
& \underline{70.77}$_{\pm 0.65}$
& \cellcolor{blue!10}\textbf{71.52}$_{\pm 0.17}$ \\
& Pass@64
& \textbf{92.08}$_{\pm 1.58}$
& \cellcolor{blue!10}\underline{91.23}$_{\pm 0.83}$ \\

\midrule
\multirow{3}{*}{GSM8K}
& Avg@64
& \textbf{81.57}$_{\pm 0.10}$
& \cellcolor{blue!10}\underline{81.43}$_{\pm 0.04}$ \\
& Pass@8
& \textbf{95.37}$_{\pm 0.08}$
& \cellcolor{blue!10}\underline{95.33}$_{\pm 0.15}$ \\
& Pass@64
& \textbf{99.12}$_{\pm 0.15}$
& \cellcolor{blue!10}\underline{98.96}$_{\pm 0.08}$ \\

\midrule
\multirow{3}{*}{MATH500}
& Avg@64
& \underline{64.43}$_{\pm 0.11}$
& \cellcolor{blue!10}\textbf{64.58}$_{\pm 0.15}$ \\
& Pass@8
& \underline{85.98}$_{\pm 0.04}$
& \cellcolor{blue!10}\textbf{86.06}$_{\pm 0.08}$ \\
& Pass@64
& \underline{93.79}$_{\pm 0.14}$
& \cellcolor{blue!10}\textbf{94.36}$_{\pm 0.22}$ \\

\midrule
\multirow{3}{*}{Minerva}
& Avg@64
& \underline{24.81}$_{\pm 0.04}$
& \cellcolor{blue!10}\textbf{24.90}$_{\pm 0.11}$ \\
& Pass@8
& \textbf{44.23}$_{\pm 0.11}$
& \cellcolor{blue!10}\underline{44.12}$_{\pm 0.28}$ \\
& Pass@64
& \underline{55.96}$_{\pm 0.28}$
& \cellcolor{blue!10}\textbf{56.92}$_{\pm 0.23}$ \\

\midrule
\multirow{3}{*}{\textbf{Average}}
& Avg@64
& \underline{36.72}$_{\pm 0.09}$
& \cellcolor{blue!10}\textbf{36.80}$_{\pm 0.06}$ \\
& Pass@8
& \underline{55.63}$_{\pm 0.19}$
& \cellcolor{blue!10}\textbf{55.67}$_{\pm 0.19}$ \\
& Pass@64
& \underline{68.26}$_{\pm 0.48}$
& \cellcolor{blue!10}\textbf{69.66}$_{\pm 0.57}$ \\

\bottomrule
\end{tabular}%
\end{table*}

\FloatBarrier
\section{Results: Additional Model Families} \label{app:extra_models}
We extend our evaluation of Entropy-Aware Mixing to LFM2.5-350M~\citep{liquidai2025lfm2,liquidAI2026350M} and Granite-3.2-2B-Instruct~\citep{ibm_granite_33} to test whether the online distillation gains observed with Qwen3 transfer across model families and student scales. 

\FloatBarrier
\subsection{LFM2.5-350M} \label{app:lfm}
We use LFM2.5-350M~\citep{liquidAI2026350M} as the student and LFM2.5-1.2B-Instruct~\citep{liquidai2025lfm2} as the teacher, training under the same setup as the main experiments. 

As shown in Table~\ref{tab:lfm_kd_online}, GeoMix-FKL achieves the strongest aggregate mathematical performance, with 14.18 Avg@64, 30.25 Pass@8, and 45.83 Pass@64, showing that entropy-aware mixing remains effective with a substantially smaller student.

\begin{table*}[h]
\centering
\caption{Online distillation results on LFM2.5-350M using MATH. Results are reported as mean$_{\pm\text{std}}$ over three seeds. \textbf{Bold} indicates the best result and \underline{underlining} the second-best result within each setting, with ties formatted identically.}
\label{tab:lfm_kd_online}
\resizebox{\textwidth}{!}{%
\begin{tabular}{llccccccc}
\toprule
\multirow{2}{*}{\textbf{Benchmark}} &
\multirow{2}{*}{\textbf{Metric}} &
\multicolumn{7}{c}{\textbf{LFM2.5-350M} (MATH)} \\
\cmidrule(lr){3-9}
&
& \textbf{FKL}
& \textbf{OPD}
& \textbf{EOPD}
& \textbf{Skew-KL}
& \textbf{TAID}
& \cellcolor{blue!10}\textbf{ConMix-FKL}
& \cellcolor{blue!10}\textbf{GeoMix-FKL} \\
\midrule

\multicolumn{9}{l}{\textbf{\textit{Mathematical reasoning}}} \\
\midrule
\multirow{3}{*}{AIME24}
& Avg@64
& 0.18$_{\pm 0.12}$
& \textbf{0.22}$_{\pm 0.13}$
& 0.13$_{\pm 0.07}$
& 0.17$_{\pm 0.11}$
& 0.15$_{\pm 0.04}$
& \cellcolor{blue!10}0.16$_{\pm 0.05}$
& \cellcolor{blue!10}\underline{0.21}$_{\pm 0.07}$ \\

& Pass@8
& 1.40$_{\pm 0.89}$
& \textbf{1.65}$_{\pm 0.98}$
& 1.00$_{\pm 0.50}$
& 1.34$_{\pm 0.77}$
& 1.18$_{\pm 0.32}$
& \cellcolor{blue!10}1.20$_{\pm 0.32}$
& \cellcolor{blue!10}\underline{1.62}$_{\pm 0.51}$ \\

& Pass@64
& 7.78$_{\pm 3.85}$
& 8.33$_{\pm 2.89}$
& 6.11$_{\pm 3.47}$
& 8.89$_{\pm 3.47}$
& \underline{9.44}$_{\pm 2.55}$
& \cellcolor{blue!10}7.22$_{\pm 0.96}$
& \cellcolor{blue!10}\textbf{10.00}$_{\pm 1.67}$ \\

\midrule
\multirow{3}{*}{AIME25}
& Avg@64
& \underline{0.21}$_{\pm 0.03}$
& \textbf{0.24}$_{\pm 0.03}$
& 0.07$_{\pm 0.03}$
& \underline{0.21}$_{\pm 0.12}$
& 0.09$_{\pm 0.08}$
& \cellcolor{blue!10}0.12$_{\pm 0.08}$
& \cellcolor{blue!10}0.10$_{\pm 0.05}$ \\

& Pass@8
& 1.55$_{\pm 0.18}$
& \textbf{1.85}$_{\pm 0.28}$
& 0.56$_{\pm 0.24}$
& \underline{1.58}$_{\pm 0.86}$
& 0.69$_{\pm 0.59}$
& \cellcolor{blue!10}0.96$_{\pm 0.61}$
& \cellcolor{blue!10}0.80$_{\pm 0.39}$ \\

& Pass@64
& 7.22$_{\pm 0.96}$
& \textbf{10.00}$_{\pm 3.33}$
& 4.44$_{\pm 1.92}$
& \underline{7.78}$_{\pm 1.92}$
& 5.00$_{\pm 4.41}$
& \cellcolor{blue!10}6.67$_{\pm 3.33}$
& \cellcolor{blue!10}4.44$_{\pm 1.92}$ \\

\midrule
\multirow{3}{*}{AMC23}
& Avg@64
& 8.90$_{\pm 0.07}$
& 10.32$_{\pm 1.04}$
& 9.64$_{\pm 0.65}$
& 11.26$_{\pm 0.59}$
& \underline{11.86}$_{\pm 0.19}$
& \cellcolor{blue!10}11.37$_{\pm 0.55}$
& \cellcolor{blue!10}\textbf{13.08}$_{\pm 0.66}$ \\

& Pass@8
& 33.06$_{\pm 0.12}$
& 36.27$_{\pm 2.19}$
& 33.82$_{\pm 2.97}$
& 35.84$_{\pm 1.47}$
& \underline{36.71}$_{\pm 0.22}$
& \cellcolor{blue!10}35.89$_{\pm 2.21}$
& \cellcolor{blue!10}\textbf{38.94}$_{\pm 0.40}$ \\

& Pass@64
& 62.92$_{\pm 5.91}$
& \textbf{67.08}$_{\pm 0.72}$
& 62.92$_{\pm 3.61}$
& 62.50$_{\pm 2.50}$
& 64.17$_{\pm 3.82}$
& \cellcolor{blue!10}59.58$_{\pm 5.05}$
& \cellcolor{blue!10}\underline{66.67}$_{\pm 2.60}$ \\

\midrule
\multirow{3}{*}{GSM8K}
& Avg@64
& 33.20$_{\pm 0.08}$
& 29.58$_{\pm 0.22}$
& 31.31$_{\pm 0.37}$
& 34.98$_{\pm 0.11}$
& \underline{36.74}$_{\pm 0.03}$
& \cellcolor{blue!10}36.73$_{\pm 0.20}$
& \cellcolor{blue!10}\textbf{38.37}$_{\pm 0.16}$ \\

& Pass@8
& 64.80$_{\pm 0.28}$
& 63.88$_{\pm 0.48}$
& 64.20$_{\pm 0.43}$
& 67.43$_{\pm 0.35}$
& 68.34$_{\pm 0.05}$
& \cellcolor{blue!10}\underline{68.56}$_{\pm 0.16}$
& \cellcolor{blue!10}\textbf{70.19}$_{\pm 0.18}$ \\

& Pass@64
& 85.00$_{\pm 0.53}$
& 86.57$_{\pm 0.75}$
& 85.00$_{\pm 0.69}$
& 86.43$_{\pm 0.67}$
& \underline{86.93}$_{\pm 0.40}$
& \cellcolor{blue!10}86.73$_{\pm 0.85}$
& \cellcolor{blue!10}\textbf{88.87}$_{\pm 0.78}$ \\

\midrule
\multirow{3}{*}{MATH500}
& Avg@64
& 21.34$_{\pm 0.07}$
& 17.90$_{\pm 0.32}$
& 19.42$_{\pm 0.20}$
& 23.97$_{\pm 0.08}$
& 24.32$_{\pm 0.11}$
& \cellcolor{blue!10}\underline{24.89}$_{\pm 0.13}$
& \cellcolor{blue!10}\textbf{26.58}$_{\pm 0.15}$ \\

& Pass@8
& 44.73$_{\pm 0.34}$
& 45.49$_{\pm 0.40}$
& 43.69$_{\pm 0.60}$
& 49.23$_{\pm 0.20}$
& 48.51$_{\pm 0.15}$
& \cellcolor{blue!10}\underline{50.67}$_{\pm 0.32}$
& \cellcolor{blue!10}\textbf{52.88}$_{\pm 0.35}$ \\

& Pass@64
& 67.13$_{\pm 0.59}$
& \underline{72.20}$_{\pm 1.05}$
& 67.27$_{\pm 1.30}$
& 70.60$_{\pm 0.26}$
& 69.80$_{\pm 0.40}$
& \cellcolor{blue!10}72.03$_{\pm 0.93}$
& \cellcolor{blue!10}\textbf{73.77}$_{\pm 1.04}$ \\

\midrule
\multirow{3}{*}{Minerva}
& Avg@64
& 4.64$_{\pm 0.13}$
& 4.27$_{\pm 0.06}$
& 4.05$_{\pm 0.16}$
& 5.42$_{\pm 0.09}$
& 5.57$_{\pm 0.04}$
& \cellcolor{blue!10}\underline{5.92}$_{\pm 0.03}$
& \cellcolor{blue!10}\textbf{6.74}$_{\pm 0.15}$ \\

& Pass@8
& 12.70$_{\pm 0.26}$
& 14.61$_{\pm 0.17}$
& 11.86$_{\pm 0.20}$
& 14.61$_{\pm 0.23}$
& 14.52$_{\pm 0.25}$
& \cellcolor{blue!10}\underline{15.19}$_{\pm 0.22}$
& \cellcolor{blue!10}\textbf{17.06}$_{\pm 0.13}$ \\

& Pass@64
& 24.63$_{\pm 1.46}$
& 28.12$_{\pm 0.80}$
& 22.18$_{\pm 0.70}$
& 27.14$_{\pm 0.65}$
& 27.70$_{\pm 1.49}$
& \cellcolor{blue!10}\underline{28.25}$_{\pm 1.18}$
& \cellcolor{blue!10}\textbf{31.25}$_{\pm 0.32}$ \\

\midrule
\multirow{3}{*}{\textbf{Average}}
& Avg@64
& 11.41$_{\pm 0.03}$
& 10.42$_{\pm 0.23}$
& 10.77$_{\pm 0.18}$
& 12.67$_{\pm 0.12}$
& 13.12$_{\pm 0.06}$
& \cellcolor{blue!10}\underline{13.20}$_{\pm 0.13}$
& \cellcolor{blue!10}\textbf{14.18}$_{\pm 0.15}$ \\

& Pass@8
& 26.37$_{\pm 0.27}$
& 27.29$_{\pm 0.27}$
& 25.85$_{\pm 0.78}$
& 28.34$_{\pm 0.40}$
& 28.32$_{\pm 0.08}$
& \cellcolor{blue!10}\underline{28.75}$_{\pm 0.49}$
& \cellcolor{blue!10}\textbf{30.25}$_{\pm 0.14}$ \\

& Pass@64
& 42.45$_{\pm 1.02}$
& \underline{45.38}$_{\pm 1.09}$
& 41.32$_{\pm 1.59}$
& 43.89$_{\pm 0.20}$
& 43.84$_{\pm 0.64}$
& \cellcolor{blue!10}43.41$_{\pm 1.36}$
& \cellcolor{blue!10}\textbf{45.83}$_{\pm 0.49}$ \\





\bottomrule
\end{tabular}%
}
\end{table*}

\FloatBarrier
\subsection{Granite-3.2-2B-Instruct} \label{app:granite}
We additionally evaluate Granite-3.2-2B-Instruct~\citep{ibm_granite_33} as the student and Granite-3.3-8B-Instruct~\citep{ibm_granite_33} as the teacher. Given that their tokenisers differ slightly, we remove four teacher-only special tokens to obtain a shared vocabulary of $49{,}155$ tokens (i.e., \texttt{<|start\_of\_cite|>}, \texttt{<|end\_of\_cite|>}, \texttt{<|start\_of\_plugin|>}, and \texttt{<|end\_of\_plugin|>}).

We use an instruction-tuned student and train on DAPO~\citep{yu2025dapo}, which better matches this post-trained setting than MATH. We otherwise follow the setup in Section~\ref{sec:experiment}, with hyperparameters in Appendix~\ref{app:hyper_params}.

Table~\ref{tab:granite_kd_online} shows that entropy-aware mixing remains competitive in this setting, with GeoMix-FKL achieving the strongest aggregate Avg@64 and Pass@8 while remaining competitive on Pass@64.

\begin{table*}[h]
\centering
\caption{Online distillation results on Granite-3.2-2B-Instruct using DAPO-Math-14k. Results are reported as mean$_{\pm\text{std}}$ over three seeds. \textbf{Bold} indicates the best result and \underline{underlining} the second-best result within each setting, with ties formatted identically.}
\label{tab:granite_kd_online}
\resizebox{\textwidth}{!}{%
\begin{tabular}{llccccccc}
\toprule
\multirow{2}{*}{\textbf{Benchmark}} &
\multirow{2}{*}{\textbf{Metric}} &
\multicolumn{7}{c}{\textbf{Granite-3.2-2B-Instruct} (DAPO-Math-14k)} \\
\cmidrule(lr){3-9}
&
& \textbf{FKL}
& \textbf{OPD}
& \textbf{EOPD}
& \textbf{Skew-KL}
& \textbf{TAID}
& \cellcolor{blue!10}\textbf{ConMix-FKL}
& \cellcolor{blue!10}\textbf{GeoMix-FKL} \\
\midrule

\multicolumn{9}{l}{\textbf{\textit{Mathematical reasoning}}} \\
\midrule

\multirow{3}{*}{AIME24}
& Avg@64
& 0.84$_{\pm 0.14}$
& \textbf{1.32}$_{\pm 0.20}$
& 1.22$_{\pm 0.12}$
& 0.97$_{\pm 0.21}$
& 1.28$_{\pm 0.21}$
& \cellcolor{blue!10}1.20$_{\pm 0.24}$
& \cellcolor{blue!10}\underline{1.30}$_{\pm 0.14}$ \\

& Pass@8
& 5.66$_{\pm 0.66}$
& 7.38$_{\pm 0.28}$
& 7.61$_{\pm 0.65}$
& 6.49$_{\pm 0.90}$
& \textbf{7.67}$_{\pm 1.37}$
& \cellcolor{blue!10}\underline{7.62}$_{\pm 0.93}$
& \cellcolor{blue!10}7.57$_{\pm 1.29}$ \\

& Pass@64
& \underline{21.11}$_{\pm 3.85}$
& \underline{21.11}$_{\pm 3.85}$
& \textbf{23.33}$_{\pm 0.00}$
& \textbf{23.33}$_{\pm 3.34}$
& \textbf{23.33}$_{\pm 5.77}$
& \cellcolor{blue!10}\textbf{23.33}$_{\pm 0.00}$
& \cellcolor{blue!10}\underline{21.11}$_{\pm 8.39}$ \\

\midrule

\multirow{3}{*}{AIME25}
& Avg@64
& 0.26$_{\pm 0.14}$
& \textbf{0.53}$_{\pm 0.24}$
& 0.40$_{\pm 0.17}$
& 0.18$_{\pm 0.03}$
& \underline{0.42}$_{\pm 0.14}$
& \cellcolor{blue!10}0.38$_{\pm 0.06}$
& \cellcolor{blue!10}0.29$_{\pm 0.12}$ \\

& Pass@8
& 2.05$_{\pm 1.10}$
& \textbf{3.70}$_{\pm 1.78}$
& 3.03$_{\pm 1.23}$
& 1.39$_{\pm 0.24}$
& \underline{3.21}$_{\pm 1.04}$
& \cellcolor{blue!10}3.03$_{\pm 0.46}$
& \cellcolor{blue!10}2.28$_{\pm 0.90}$ \\

& Pass@64
& 14.44$_{\pm 8.39}$
& \textbf{22.22}$_{\pm 8.82}$
& 17.78$_{\pm 6.94}$
& 11.11$_{\pm 1.92}$
& \underline{18.89}$_{\pm 5.09}$
& \cellcolor{blue!10}\textbf{22.22}$_{\pm 1.92}$
& \cellcolor{blue!10}18.55$_{\pm 6.94}$ \\

\midrule

\multirow{3}{*}{AMC23}
& Avg@64
& 14.06$_{\pm 0.69}$
& 14.02$_{\pm 1.11}$
& 13.35$_{\pm 0.23}$
& 13.33$_{\pm 0.63}$
& \textbf{14.83}$_{\pm 0.72}$
& \cellcolor{blue!10}14.29$_{\pm 0.67}$
& \cellcolor{blue!10}\underline{14.81}$_{\pm 0.80}$ \\

& Pass@8
& 46.11$_{\pm 3.26}$
& \underline{47.65}$_{\pm 2.09}$
& 45.33$_{\pm 2.28}$
& 45.68$_{\pm 2.89}$
& 46.90$_{\pm 1.95}$
& \cellcolor{blue!10}\underline{47.65}$_{\pm 0.98}$
& \cellcolor{blue!10}\textbf{47.87}$_{\pm 1.82}$ \\

& Pass@64
& 80.00$_{\pm 8.66}$
& \textbf{81.67}$_{\pm 2.89}$
& \textbf{81.67}$_{\pm 1.44}$
& 80.00$_{\pm 0.00}$
& 79.17$_{\pm 5.20}$
& \cellcolor{blue!10}\underline{80.83}$_{\pm 3.82}$
& \cellcolor{blue!10}\textbf{81.67}$_{\pm 6.29}$ \\

\midrule

\multirow{3}{*}{GSM8K}
& Avg@64
& 73.05$_{\pm 0.17}$
& \textbf{76.33}$_{\pm 0.15}$
& 74.91$_{\pm 0.22}$
& 73.36$_{\pm 0.19}$
& 74.24$_{\pm 0.16}$
& \cellcolor{blue!10}73.75$_{\pm 0.05}$
& \cellcolor{blue!10}\underline{76.29}$_{\pm 0.21}$ \\

& Pass@8
& 93.70$_{\pm 0.10}$
& \underline{94.20}$_{\pm 0.08}$
& 94.00$_{\pm 0.16}$
& 93.71$_{\pm 0.09}$
& 93.99$_{\pm 0.16}$
& \cellcolor{blue!10}94.06$_{\pm 0.25}$
& \cellcolor{blue!10}\textbf{94.47}$_{\pm 0.18}$ \\

& Pass@64
& 98.00$_{\pm 0.20}$
& 97.80$_{\pm 0.20}$
& \underline{98.07}$_{\pm 0.23}$
& 97.93$_{\pm 0.31}$
& 98.00$_{\pm 0.35}$
& \cellcolor{blue!10}\textbf{98.20}$_{\pm 0.35}$
& \cellcolor{blue!10}97.73$_{\pm 0.12}$ \\

\midrule

\multirow{3}{*}{MATH500}
& Avg@64
& 38.31$_{\pm 0.25}$
& \underline{41.48}$_{\pm 0.10}$
& 40.74$_{\pm 0.09}$
& 37.94$_{\pm 0.17}$
& 39.90$_{\pm 0.02}$
& \cellcolor{blue!10}39.36$_{\pm 0.28}$
& \cellcolor{blue!10}\textbf{41.80}$_{\pm 0.13}$ \\

& Pass@8
& 68.85$_{\pm 0.52}$
& 70.12$_{\pm 0.04}$
& \underline{70.20}$_{\pm 0.27}$
& 68.17$_{\pm 0.54}$
& 69.78$_{\pm 0.17}$
& \cellcolor{blue!10}69.53$_{\pm 0.42}$
& \cellcolor{blue!10}\textbf{70.58}$_{\pm 0.38}$ \\

& Pass@64
& 87.53$_{\pm 0.92}$
& \underline{87.87}$_{\pm 0.50}$
& \textbf{88.00}$_{\pm 0.20}$
& 86.60$_{\pm 0.87}$
& 86.87$_{\pm 0.23}$
& \cellcolor{blue!10}\textbf{88.00}$_{\pm 0.80}$
& \cellcolor{blue!10}\underline{87.87}$_{\pm 1.39}$ \\

\midrule

\multirow{3}{*}{Minerva}
& Avg@64
& 12.63$_{\pm 0.06}$
& \underline{14.42}$_{\pm 0.15}$
& 13.81$_{\pm 0.09}$
& 12.78$_{\pm 0.17}$
& 13.48$_{\pm 0.09}$
& \cellcolor{blue!10}13.25$_{\pm 0.10}$
& \cellcolor{blue!10}\textbf{14.47}$_{\pm 0.15}$ \\

& Pass@8
& 31.49$_{\pm 0.60}$
& \underline{33.35}$_{\pm 0.24}$
& 32.97$_{\pm 0.38}$
& 31.46$_{\pm 0.59}$
& 32.70$_{\pm 0.38}$
& \cellcolor{blue!10}32.10$_{\pm 0.03}$
& \cellcolor{blue!10}\textbf{33.79}$_{\pm 0.55}$ \\

& Pass@64
& 47.67$_{\pm 2.03}$
& 47.67$_{\pm 1.12}$
& \textbf{49.51}$_{\pm 1.29}$
& 47.18$_{\pm 1.39}$
& 48.65$_{\pm 0.85}$
& \cellcolor{blue!10}47.18$_{\pm 1.39}$
& \cellcolor{blue!10}\underline{49.14}$_{\pm 2.25}$ \\

\midrule

\multirow{3}{*}{\textbf{Average}}
& Avg@64
& 23.19$_{\pm 0.24}$
& \underline{24.68}$_{\pm 0.33}$
& 24.07$_{\pm 0.15}$
& 23.09$_{\pm 0.23}$
& 24.03$_{\pm 0.22}$
& \cellcolor{blue!10}23.70$_{\pm 0.23}$
& \cellcolor{blue!10}\textbf{24.83}$_{\pm 0.26}$ \\

& Pass@8
& 41.31$_{\pm 1.04}$
& \underline{42.73}$_{\pm 0.75}$
& 42.19$_{\pm 0.83}$
& 41.15$_{\pm 0.87}$
& 42.38$_{\pm 0.85}$
& \cellcolor{blue!10}42.33$_{\pm 0.51}$
& \cellcolor{blue!10}\textbf{42.76}$_{\pm 0.85}$ \\

& Pass@64
& 58.13$_{\pm 4.01}$
& 59.72$_{\pm 2.90}$
& \underline{59.73}$_{\pm 1.68}$
& 57.69$_{\pm 1.30}$
& 59.15$_{\pm 2.92}$
& \cellcolor{blue!10}\textbf{59.96}$_{\pm 1.38}$
& \cellcolor{blue!10}59.35$_{\pm 4.23}$ \\





\bottomrule
\end{tabular}%
}
\end{table*}

\end{document}